\documentclass{article} 
\pdfoutput=1 
\usepackage[dvipsnames]{xcolor}
\usepackage{natbib}
\usepackage{iclr2019_conference,times}

\usepackage{amssymb,amsmath}
\usepackage{amsthm}

\usepackage{amsmath,amsfonts,bm}

\def\figref#1{figure~\ref{#1}}
\def\Figref#1{Figure~\ref{#1}}
\def\twofigref#1#2{figures \ref{#1} and \ref{#2}}

\def\secref#1{section~\ref{#1}}

\def\eqref#1{equation~\ref{#1}}

\def\eqrefp#1{equation~(\ref{#1})} 

\def\thmref#1{theorem~\ref{#1}}
\def\tableref#1{Table~\ref{#1}}  

\def\1{\bm{1}}

\def\eps{{\epsilon}}

\DeclareMathAlphabet{\mathsfit}{\encodingdefault}{\sfdefault}{m}{sl}
\SetMathAlphabet{\mathsfit}{bold}{\encodingdefault}{\sfdefault}{bx}{n}

\newcommand{\Var}{\mathrm{Var}}

\newcommand{\diag}{\mathbf{diag}}

\DeclareMathOperator{\Tr}{Tr}

\definecolor{mydarkblue}{rgb}{0,0.08,0.45}
\usepackage{hyperref}
\hypersetup{colorlinks,linkcolor={mydarkblue},citecolor={mydarkblue},urlcolor={mydarkblue}}
\usepackage{url}

\usepackage{booktabs} 

\usepackage{tikz}
\usetikzlibrary{bayesnet}
\usepackage{caption, multirow,multicol}
\usepackage[labelformat=simple]{subcaption}

\newtheorem{theorem}{Theorem}[section]

\newcommand{\explainEq}[1]{\overset{\underset{\mathrm{(#1)}}{}}{=}}
\newcommand{\explainLeq}[1]{\overset{\underset{\mathrm{(#1)}}{}}{\leq}}

\theoremstyle{definition}
\newtheorem{definition}{Definition}

\newtheorem{proposition}{Proposition}

\title{Distributed generation of privacy preserving data with user customization}

\author{Xiao Chen, Thomas Navidi, Stefano Ermon,  Ram Rajagopal \\
Stanford University\\
\texttt{\{markcx@, tnavidi@, ermon@cs., ramr@\}stanford.edu}
}

\iclrfinalcopy 
\begin{document}

\maketitle

\begin{abstract}
    Distributed devices such as mobile phones can produce and store large amounts of data that can enhance machine learning models; however, this data may contain private information specific to the data owner that prevents the release of the data. We wish to reduce the correlation between user-specific private information and data while maintaining the useful information. Rather than learning a large model to achieve privatization from end to end, we introduce a decoupling of the creation of a latent representation and the privatization of data that allows user-specific privatization to occur in a distributed setting with limited computation and minimal disturbance on the utility of the data. We leverage a Variational Autoencoder (VAE) to create a compact latent representation of the data; however, the VAE remains fixed for all devices and all possible private labels. We then train a small generative filter to perturb the latent representation based on individual preferences regarding the private and utility information. The small filter is trained by utilizing a GAN-type robust optimization that can take place on a distributed device. We conduct experiments on three popular datasets: MNIST, UCI-Adult, and CelebA, and give a thorough evaluation including visualizing the geometry of the latent embeddings and estimating the empirical mutual information to show the effectiveness of our approach.     
\end{abstract}

\section{Introduction}
%
The success of machine learning algorithms relies on not only technical methodologies, but also the availability of large datasets such as images  \citep{krizhevsky2012imagenet, oshri2018infrastructure}; however, data can often contain sensitive information, such as race or age, that may hinder the owner's ability to release the data to exploit its utility. We are interested in exploring methods of providing privatized data such that sensitive information cannot be easily inferred from the adversarial perspective, while preserving the utility of the dataset. In particular, we consider a setting where many participants are independently gathering data that will be collected by a third party. Each participant is incentivized to label their own data with useful information; however, they have the option to create private labels for information that they do not wish to share with the database. In the case where data contains a large amount of information such as images, there can be an overwhelming number of potential private and utility label combinations (skin color, age, gender, race, location, medical conditions, etc.). The large number of combinations prevents training a separate method to obscure each set of labels at a central location. Furthermore, when participants are collecting data on their personal devices such as mobile phones, they would like to remove private information before the data leaves their devices. Both the large number of personal label combinations coupled with the use of mobile devices requires a privacy scheme to be computationally efficient on a small device. In this paper, we propose a method of generating private datasets that makes use of a fixed encoding, thus requiring only a few small neural networks to be trained for each label combination. This approach allows data collecting participants to select any combination of private and utility labels and remove them from the data on their own mobile device before sending any information to a third party.

In the context of publishing datasets with privacy and utility guarantees, we briefly review a number of similar approaches that have been recently considered, and discuss why they are inadequate at performing in the distributed and customizable setting we have proposed. Traditional work in generating private datasets has made use of differential privacy (DP)\citep{dwork2006calibrating}, which involves injecting certain random noise into the data to prevent the identification of sensitive information \citep{dwork2006calibrating, dwork2011differential, dwork2014algorithmic}. However, finding a globally optimal perturbation using DP may be too stringent of a privacy condition in many high dimensional data applications. Therefore, more recent literature describes research that commonly uses Autoencoders \citep{kingma2013auto} to create a compact latent representation of the data, which does not contain private information, but does encode the useful information \citep{edwards2015censoring, abadi2016learning, beutel2017data, madras2018learning, song2018learning, chen2018understanding}. A few papers combine strategies involving both DP and Autoencoders \citep{hamm2017minimax, liu2017deeprotect}; however, all of these recent strategies require training a separate Autoencoder for each possible combination of private and utility labels. Training an Autoencoder for each privacy combination can be computationally prohibitive, especially when working with high dimensional data or when computation must be done on a small local device such as a mobile phone. Therefore, such methods are unable to handle our proposed scenario where each participant must locally train a data generator that obscures their individual choice of private and utility labels. Additionally, reducing the computation and communication burden when dealing with distributed data is beneficial in many other potential applications such as federated learning \citep{mcmahan2016communication}.

\begin{figure*}[!hbpt]
    \centerline{
    \begin{subfigure}[t]{0.49\columnwidth}
    \includegraphics[width=1\textwidth]{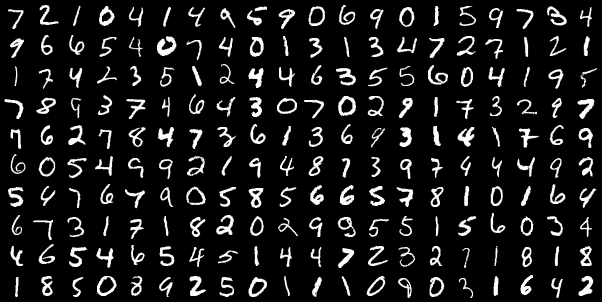}
    \caption{Sample of original images}
    \end{subfigure}
    \begin{subfigure}[t]{0.49\columnwidth}
    \includegraphics[width=1\textwidth]{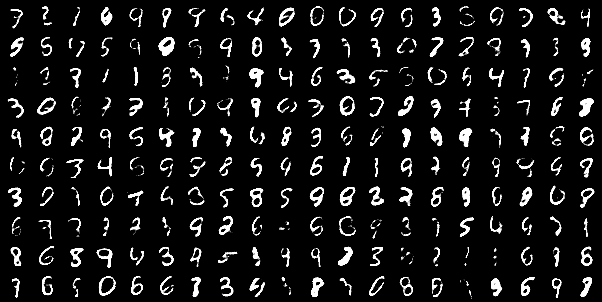}
    \caption{Same images perturbed to privatize digit ID}
    \end{subfigure}
    }
    \caption{Visualization of digits pre- and post-noise injection and adversarial training. We find that digit IDs are randomly switched while circle digits remain circle digits and non-circle digits remain as non-circle digits. }
    \label{fig:sample_MNIST}
\end{figure*}

We verify our idea on three datasets. The first is the {MNIST} dataset \citep{lecun-mnisthandwrittendigit-2010} of handwritten digits, commonly used as a synthetic example in machine learning literature. We have two cases involving this dataset: In \textbf{MNIST Case 1}, we preserve information regarding whether the digit contains a circle (i.e. digits \textsf{0,6,8,9}), but privatize the value of the digit itself. In \textbf{MNIST Case 2}, we preserve information on the parity (even or odd digit), but privatize whether or not the digit is greater than or equal to 5. \Figref{fig:sample_MNIST} shows a sample of the original dataset along with the same sample now perturbed to remove information on the digit identity, but maintain the digit as a circle containing digit. The input to the algorithm is the original dataset with labels, while the output is the perturbed data as shown. The second is the UCI-adult income dataset \citep{Dua:2019} that has 45222 anonymous adults from the 1994 US Census to predict whether an individual has an annual income over {\$50,000} or not. The third dataset is the CelebA dataset \citep{liu2015faceattributes} containing color images of celebrity faces. For this realistic example, we preserve whether the celebrity is smiling, while privatizing many different labels (gender, age, etc.) independently to demonstrate our capability to privatize a wide variety of labels with a single latent representation.

Primarily, this paper introduces a decoupling of the creation of a latent representation and the privatization of data that allows the privatization to occur in a distributed setting with limited computation and minimal disturbance on the utility of the data. Additional contributions include a variant on the Autoencoder to improve robustness of the decoder for reconstructing perturbed data, and thorough investigations into: (i) the latent geometry of the data distribution before and after privatization, (ii) how training against a cross entropy loss adversary impacts the mutual information between the data and the private label, and (iii) the use of \emph{f}-divergence based constraints in a robust optimization and their relationship to more common norm-ball based constraints and differential privacy. These supplemental investigations provide a deeper understanding of how our scheme is able to achieve privatization.

%
%
\section{Problem Statement and Methodology}
Inspired by a few recent studies \citep{louizos2015variational, huang2017context, madras2018learning, chen2018understanding}, we consider the data privatization as a game between two players: the data generator (data owner) and the adversary (discriminator). The generator tries to inject noise that will privatize certain sensitive information contained in the data, while the adversary tries to infer this sensitive information from the data. In order to deal with high dimensional data, we first learn a latent representation or manifold of the data distribution, and then inject noise with specific latent features to reduce the correlation between released data and sensitive information. After the noise has been added to the latent vector, the data can be reconstructed and published without fear of releasing sensitive information. To summarize, the input to our system is the original dataset with both useful and private labels, and the output is a perturbed dataset that has reduced statistical correlation with the private labels, but has maintained information related to the useful labels.

\begin{figure*}[!hpbt]
    \centerline{
    \includegraphics[width=0.74\textwidth]{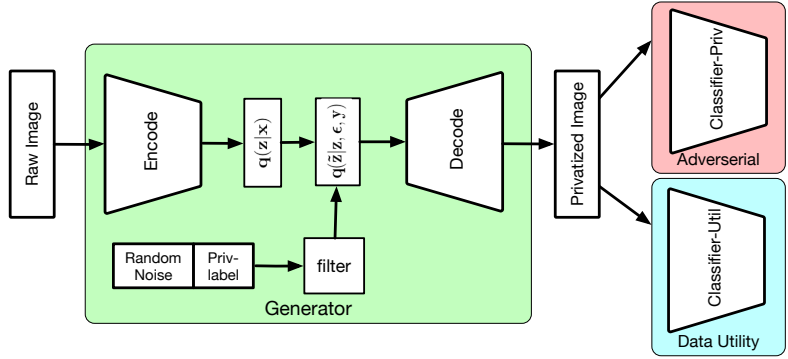}
    }
    \caption{Privatization architecture. We decompose the whole privatization procedure into two steps: 1) training an encoder and decoder; 2) learning a generative filter. }
    \label{fig:learn:schematic:architecture}
\end{figure*}

We consider the general setting where a data owner holds a dataset $\mathcal{D}$ that consists of original data $X$, private/sensitive labels $Y$, and useful labels $U$. Thus, each sample $i$ has a record $(x_i, y_i, u_i) \in \mathcal{D}$. We denote the function $g$ as a general mechanism for the data owner to release the data. The released data is denoted as $\{\tilde{X}, \tilde{U}\}$. Because $Y$ is private information, it won't be released. Thus, for the record $i$, the released data can be described as $(\tilde{x}_i, \tilde{u}_i) = g(x_i, y_i, u_i).$ We simplify the problem by considering only the case that $\tilde{x}_i = g(x_i, y_i)$\footnote{We maintain the $U$ to be unchanged} for the following description. The corresponding perturbed data $\tilde{X} = g(X, Y)$ and utility attributes $U$ are published for use. The adversary builds a learning algorithm $h$ to infer the sensitive information given the released data, i.e. $ \hat{Y} = h(\tilde{X})$ where $\hat{Y}$ is the estimate of $Y$. The goal of the adversary is to minimize the inference loss $\ell{(\hat{Y}, Y)} = \ell{\Big(h\big(g(X, Y)\big), Y\Big)}$ on the private labels. Similarly, we denote the estimate of utility labels as $\hat{U} = \nu(\tilde{X}) = \nu(g(X, Y)) $. We quantify the utility of the released data through another loss function $\tilde{\ell}$ that captures the utility attributes, i.e. $\tilde{\ell}(\tilde{X}, U ) = \tilde{\ell}\Big(\nu\big(g(X, Y)\big), U \Big)$. The data owner wants to maximize the loss that the adversary experiences to protect the sensitive information while maintaining the data utility by minimizing the utility loss. Given the previous settings, the data-releasing game can be expressed as follows:
\begin{align*}
    \max_{g \in \mathcal{G}}\Big\{ \min_{h \in \mathcal{H}}\mathbb{E}\Big[ \ell{\Big(h\big(g(X, Y)\big), Y\Big)}\Big]  - \beta \min_{\nu \in \mathcal{V}}  \mathbb{E}\Big[\tilde{\ell}\Big(\nu\big( g(X, Y) \big), U \Big)\Big]  \Big\},
\end{align*}
where $\beta$ is a hyper-parameter weighing the trade-off between different losses, and the expectation is taken over all samples from the dataset. The loss functions in this game are flexible and can be tailored to a specific metric that the data owner is interested in. For example, a typical loss function for classification problems is cross-entropy loss \citep{de2005tutorial}. Because optimizing over the functions $g, h, \nu$ is hard to implement, we use a variant of the min-max game that leverages neural networks to approximate the functions. The foundation of our approach is to construct a good posterior approximation for the data distribution in latent space $\mathcal{Z}$, and then to inject context aware noise through a filter in the latent space, and finally to run the adversarial training to achieve convergence, as illustrated in
figure~\ref{fig:learn:schematic:architecture}. Specifically, we consider the data owner playing the generator role that comprises a Variational Autoencoder (VAE) \citep{kingma2013auto} 
structure with an additional noise injection filter in the latent space. We use $\theta_g$, $\theta_h$, and $\theta_{\nu}$ to denote the parameters of neural networks that represent the data owner (generator), adversary (discriminator), and utility learner (util-classifier), respectively. Moreover, the parameters of generator $\theta_g$ consists of the encoder parameters $\theta_e$, decoder parameters $\theta_d$, and filter parameters $\theta_f$. The encoder and decoder parameters are trained independently of the privatization process and left fixed. Hence, we have 
\begin{align*}
     & \max_{\theta_f}\Big\{ \min_{\theta_h}\mathbb{E}\Big[\ell{\Big(h_{\theta_h}\big(g_{\theta_g}(X, Y)\big), Y\Big)}\Big]  - \beta \min_{\theta_{\nu}} \mathbb{E}\Big[\tilde{\ell} \Big(\nu_{\theta_{\nu}}\big( g_{\theta_g}(X, Y) \big), U \Big)\Big]  \Big\} \\
     & s.t. \qquad D\Big(g_{\theta_e}(X), g_{\theta_f}\big(g_{\theta_e}(X), \eps, Y\big)\Big) \leq b ,
\end{align*}
where $D$ is a distance or divergence measure, and $b$ is the corresponding distortion budget. 

In principle, we perform the following three steps to complete each experiment.

1) \textbf{Train a VAE} for the generator without noise injection or min-max adversarial training. Rather than imposing the distortion budget at the beginning, we first train the following objective 
\begin{align}
    \min_{\theta_e, \theta_d} -\mathbb{E}_{q({z}|{x};\theta_e)} [\log p({x}|{z};\theta_d)] + D_{\text{KL}}\big(q({z}|{x};\theta_e) || p({z})\big) \label{eq:vae:nelbo}
\end{align} 
where the posterior distribution $q(z|x; \theta_e)$ is characterized by an encoding network $g_{\theta_e}(x)$, and $p(x|z; \theta_d)$ is similarly the decoding network $g_{\theta_d}(z)$. The distribution $p(z)$, is a prior distribution that is usually chosen to be a multivariate Gaussian distribution for reparameterization purposes \citep{kingma2013auto}. When dealing with high dimensional data, we develop a variant of the preceding objective that captures three items: the reconstruction loss, KL divergence on latent representations, and improved robustness of the decoder network to perturbations in the latent space (as shown in \eqrefp{eq:supp:vae:loss:variant}). We discuss more details of training a VAE and its variant in \secref{supp:tech:approch:vae:explained}.

2) \textbf{Formulate a robust optimization} to run the min-max GAN-type training \citep{goodfellow2014generative} with the noise injection, which comprises a linear or nonlinear filter \footnote{The filter can be a small neural network.}, while freezing the weights of the encoder and decoder. In this phase, we instantiate several divergence metrics and various distortion budgets to run our experiments (details in section \ref{supp:tech:approach:noise_inject:train}). When we fix the encoder, the latent variable $z=g_{\theta_e}(x)$ (or $z \sim q_{\theta_e}(z|x)$ or $q_{\theta_e}$ for short), and the new altered latent representation $\tilde{z}$ can be expressed as $\tilde{z}=g_{\theta_f}(z, \epsilon, y)$, where $g_{\theta_f}$ represents a filter function (or $\tilde{z} \sim q_{\theta_f}$ for short). The classifiers $h$ and $\nu$ can take the latent vector as input when training the filter to reduce the computational burden as is done in our experiments. We focus on a canonical form for the adversarial training and cast our problem into the following robust optimization problem: 
\begin{align}
    & \min_{\theta_{h}}\Big\{ \max_{q_{\theta_f}}\mathbb{E}_{q_{\theta_f}}\big(\ell(h_{\theta_h}; \tilde{z}) \big) - \beta\mathbb{E}_{q_{\theta_f}}\big(\tilde{\ell}(\nu_{\theta_{\nu}}; \tilde{z}) \big) \Big\}\\
    s.t. & \qquad D_f(q_{\theta_f}||q_{\theta_e}) \leq b \\
    & \qquad  \mathbb{E}q_{\theta_f} = 1 \\
    & \qquad \theta_{\nu} = \arg\min_{\theta_{\nu}}\mathbb{E}_{q_{\theta_f}}\big( \tilde{\ell}(\nu_{\theta_{\nu}}; \tilde{z}) \big),
\end{align}
where $D_f$ is the \emph{$f$-divergence} in this case. We disclose some connections between the divergence based constraints and norm-ball perturbation based constraints in \secref{supp:tech:approach:noise_inject:train}. We also discuss the ability to incorporate differential privacy mechanisms, which we use as a baseline, through the constrained optimization as well as its distinctions in \secref{supp:tech:approach:diff:privacy} of the appendix.
%

3) \textbf{Learn the adaptive classifiers} for both the adversary and utility according to $[\tilde{X}, Y, U]$ \big(or $[\tilde{Z}, Y, U]$ if the classifier takes the latent vector as input\big), where the perturbed data $\tilde{X}$ (or $\tilde{Z}$) is generated based on the trained generator. We validate the performance of our approach by comparing metrics such as classification accuracy and empirical mutual information. Furthermore, we visualize the geometry of the latent representations, e.g. \figref{fig:circle:latent:geo:embedding:image}, to give intuitions behind how our framework achieves privacy.

\section{Experiments and Results}

We present the results of experiments on the datasets MNIST, UCI-adult, and CelebA to demonstrate the effectiveness of our approach. 

\begin{figure*}[!hpbt]
    \centerline{
    \begin{subfigure}[t]{0.37\columnwidth}
    \includegraphics[width=1\textwidth]{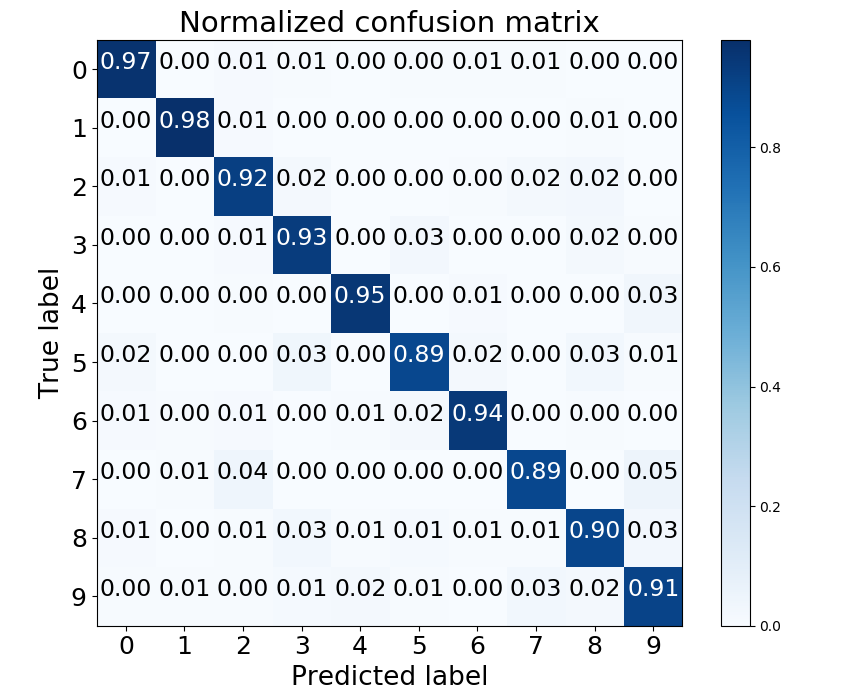}
    \caption{raw}
    \label{fig:circle:acc:confusion:mat:raw}
    \end{subfigure}
    \begin{subfigure}[t]{0.37\columnwidth}
    \includegraphics[width=1\textwidth]{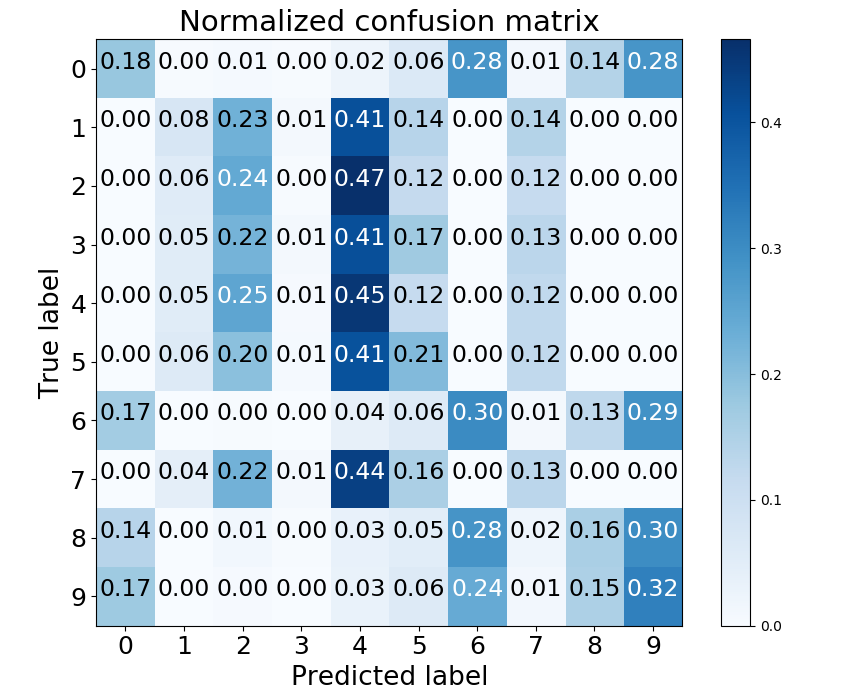}
    \caption{privatized}
    \label{fig:circle:acc:confusion:mat:perturb}
    \end{subfigure}
    \begin{subfigure}[t]{0.41\columnwidth}
    \includegraphics[width=1\textwidth]{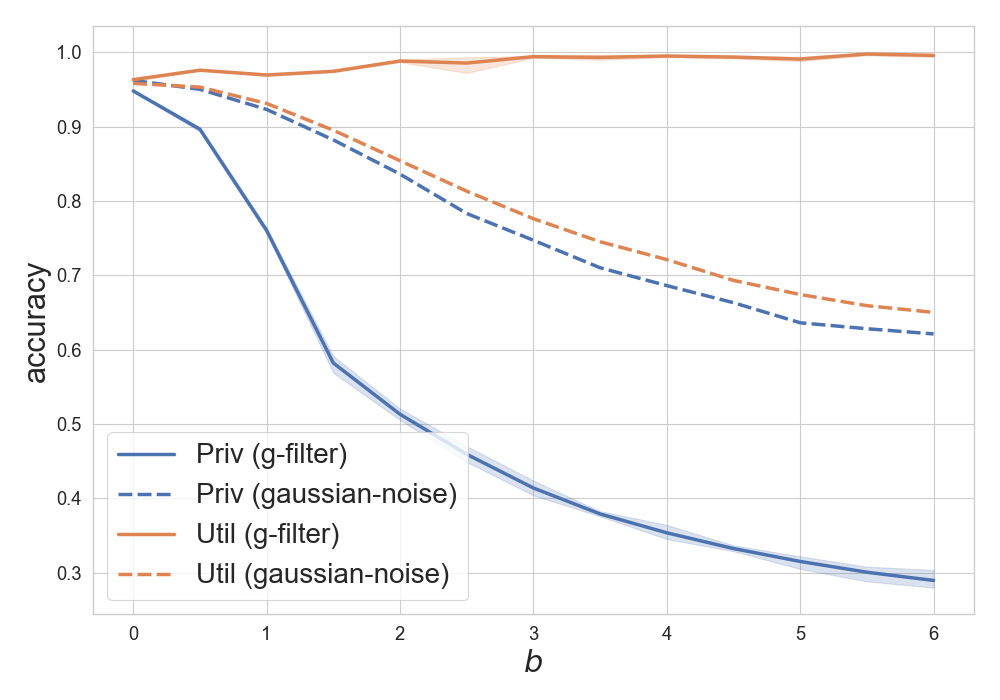}
    \caption{classification accuracy}
    \label{fig:circle:acc:distortion:compare}
    \end{subfigure}
    }
    \caption{Classifying digits in MNIST. Original digits can be easily classified with more than 90\% accuracy on average, yet the new perturbed digits have a significantly lower accuracy as expected. Specifically, many circle digits are incorrectly classified as other circle digits and similarly for the non-circle digits. \Figref{fig:circle:acc:distortion:compare} demonstrates that classification accuracy on the private label decreases quickly while classification on the utility label remains nearly constant as the distortion budget increases. Furthermore, our approach is superior to the baseline Gaussian mechanism based on differential privacy.}
    \label{fig:mnist:circle:clf:res}
\end{figure*}

\begin{figure*}[!hpbt]
    \centerline{
    \begin{subfigure}[t]{0.37\columnwidth}
    \includegraphics[width=1\textwidth]{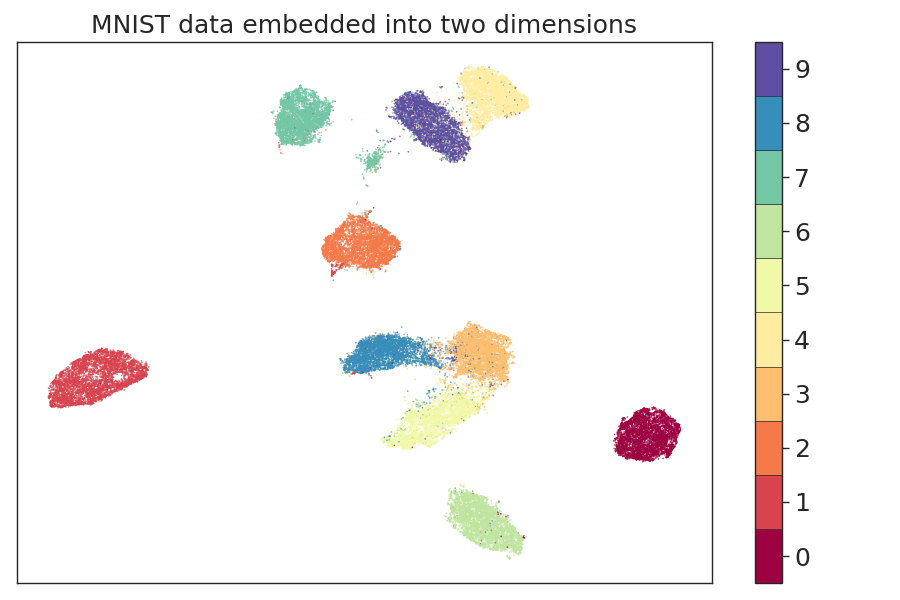}
    \caption{raw}
    \label{fig:circle:latent:geo:raw}
    \end{subfigure}
    \begin{subfigure}[t]{0.37\columnwidth}
    \includegraphics[width=1\textwidth]{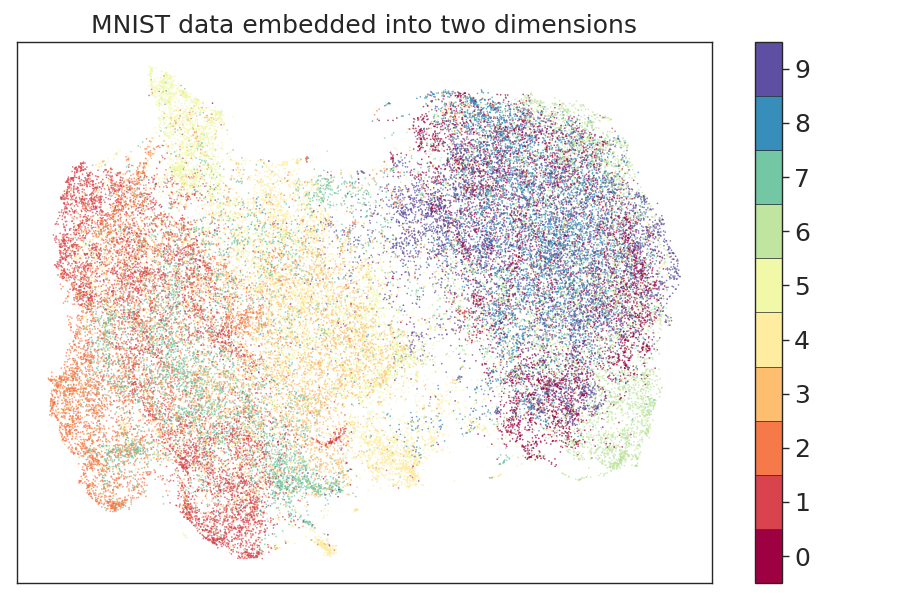}
    \caption{moderate perturbation }
    \label{fig:circle:latent:geo:moderate}
    \end{subfigure}
    \begin{subfigure}[t]{0.37\columnwidth}
    \includegraphics[width=1\textwidth]{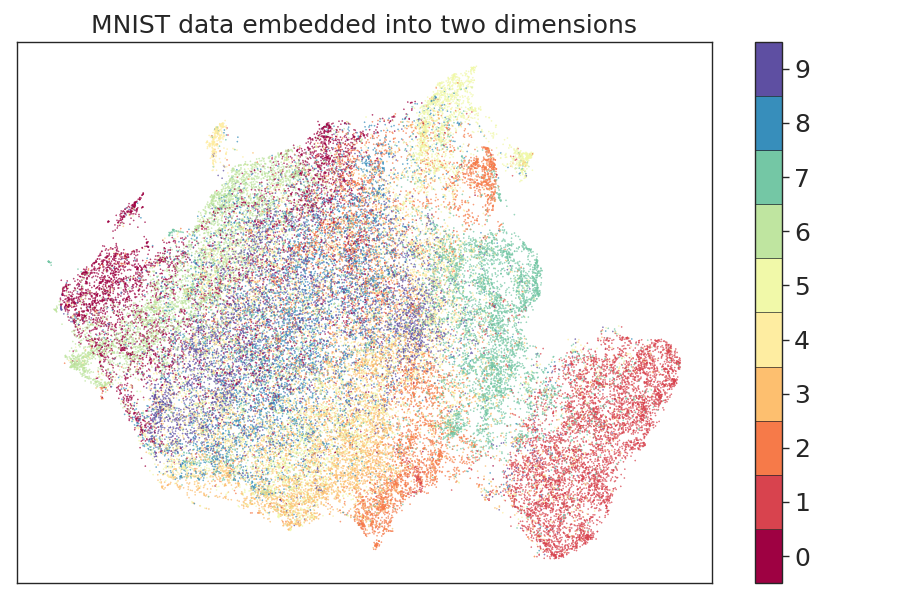}
    \caption{heavy perturbation}
    \label{fig:circle:latent:geo:heavy}
    \end{subfigure}
    }
    \caption{Visualization of the latent geometry. The original embedding in \figref{fig:circle:latent:geo:raw} is clearly segmented into individual clusters for each digit; however, when we allow a distortion budget of $b=1.5$, as shown in \figref{fig:circle:latent:geo:moderate}, the digits are separated according to the circle or non-circle property by the gap between the left and right clouds of points. A larger distortion budget nests all samples close together with some maintained local clusters as seen in \figref{fig:circle:latent:geo:heavy}. }
    \label{fig:circle:latent:geo:embedding:image}
\end{figure*}

\textbf{MNIST Case 1}: We consider the digit number itself as the private attribute and the digit containing a circle or not as the utility attribute. Figure~\ref{fig:sample_MNIST} shows samples of this case as introduced before. Specific classification results before and after privatization are given in the form of confusion matrices in \twofigref{fig:circle:acc:confusion:mat:raw}{fig:circle:acc:confusion:mat:perturb}, demonstrating a significant reduction in private label classification accuracy. These results are supported by our illustrations of the latent space geometry in \Figref{fig:circle:latent:geo:embedding:image} via uniform manifold approximation and projection (UMAP) \citep{mcinnes2018umap}. Specifically, \figref{fig:circle:latent:geo:moderate} shows a clear separation between circle digits (on the right) and non-circle digits (on the left). We also investigate the sensitivity of classification accuracy for both labels with respect to the distortion budget (for KL-divergence) in \Figref{fig:circle:acc:distortion:compare}, demonstrating that increasing the distortion budget rapidly decreases the private label accuracy while maintaining the utility label accuracy. We also compare these results to a baseline method based on differential privacy, (an additive Gaussian mechanism discussed in \secref{supp:tech:approach:diff:privacy}), and we find that this additive Gaussian mechanism performs worse than our generative adversarial filter in terms of keeping the utility and protecting the private labels because the Gaussian mechanism yields lower utility and worse privacy (i.e. higher prediction accuracy of private labels) than the min-max generative filter approach.         

\textbf{MNIST Case 2}: This case has the same setting as the experiment given in \citet{rezaei2018protecting} where we consider odd or even digits as the target utility and large or small value ($\geq 5$) as the private label. Rather than training a generator based on a fixed classifier, as done in \citet{rezaei2018protecting}, we take a different modeling and evaluation approach that allows the adversarial classifier to update dynamically. We find that the classification accuracy of the private attribute drops down from 95\% to 65\% as the distortion budget grows. Meanwhile our generative filter doesn't deteriorate the target utility too much, which maintains a classification accuracy above 87\% for the utility label as the distortion increases, as shown in \figref{fig:mnist:acc:inc:b:priv:geq5:util:oddeven}. We discuss more results in the appendix \secref{supp:mnist:more:res:w:mi}, together with results verifying the reduction of mutual information between the data and the private labels.   

While using the privatization scheme from case 2, we measure the classification accuracy of the circle attribute from case 1. This tests how the distortion budget prevents excessive loss of information of non-target attributes. The circle attribute from case 1 is not included in the loss function when training for case 2; however, as seen in \tableref{tab:mnist:non-target}, the classification accuracy on the circle is not more diminished than the target attribute (odd). A more detailed plot of the classification accuracy can be found in \figref{fig:mnist:acc:inc:b:priv:geq5:util:non-target} in the appendix \secref{supp:mnist:more:res:w:mi}. This demonstrates that the privatized data maintains utility beyond the predefined utility labels used in training.

\begin{table}[!hbpt]
    \centering
    \caption{Accuracy of private label ($\geq 5$), target label (odd), and non-target label (circle) for MNIST dataset. The raw embedding yield by VAE is denoted as \emph{emb-raw}. The embedding yield from generative filter is denoted as \emph{emb-g-filter}.}
    \label{tab:mnist:non-target}
    \begin{small}
    \begin{tabular}{c|c|c|c}
    \toprule 
    \textbf{Data} & \textbf{Private attr.} & \textbf{Utility attr.}  & \textbf{Non-target attr.} \\
    \midrule 
        emb-raw & 0.951 & 0.952 & 0.951 \\
        emb-g-filter & 0.687 & 0.899 & 0.9 \\
    \bottomrule    
    \end{tabular}
    \end{small}
\end{table}

\textbf{UCI-Adult}: We conduct the experiment by setting the private label to be gender and the utility label to be income. All the data is preprocessed to binary values for the ease of training. We compare our method with the models of Variational Fair AutoEncoder (VFAE)\citep{louizos2015variational} and Lagrangian Mutual Information-based Fair Representations (LMIFR) \citep{song2018learning} to show the performance. The corresponding accuracy and area-under receiver operating characteristic curve (AUROC) of classifying private label and utility label are shown in \tableref{tab:uciadult:benchmark:res}. Our method has the lowest accuracy and the smallest AUROC on the privatized gender attribute. Although our method doesn't perform best on classifying the utility label, it still achieves comparable results in terms of both accuracy and AUROC which are described in \tableref{tab:uciadult:benchmark:res}.           

\begin{table}[!hbpt]
    \centering
    \caption{Accuracy (acc.) and Area-Under-ROC (auroc.) of private label (gender) and target label (income) for UCI-Adult dataset.}
    \label{tab:uciadult:benchmark:res}
    \begin{small}
    \begin{tabular}{c|c|c|c|c}
    \toprule 
    \multirow{1}{*}{\textbf{Model}} & \multicolumn{2}{c|}{\textbf{Private attr. }}  & \multicolumn{2}{c}{\textbf{Utility attr. }}  \\
    & acc. & auroc. & acc. & auroc. \\ 
    \midrule 
        VAE \citep{kingma2013auto}  & 0.850 $\pm$ 0.007 & 0.843 $\pm$ 0.004 & 0.837 $\pm$ 0.009 & 0.755 $\pm$ 0.005 \\
        VFAE \citep{louizos2015variational} & 0.802 $\pm$ 0.009 & 0.703 $\pm$ 0.013 & \textbf{0.851 $\pm$ 0.004} & 0.761 $\pm$ 0.011 \\
        LMIFR \citep{song2018learning} & 0.728 $\pm$ 0.014 & 0.659 $\pm$ 0.012 & 0.829 $\pm$ 0.009 & 0.741 $\pm$ 0.013  \\
        Ours (w. generative filter) & \textbf{ 0.717 $\pm$ 0.006} & 0.632 $\pm$ 0.011 & 0.822 $\pm$ 0.005 & 0.731 $\pm$ 0.015 \\
    \bottomrule    
    \end{tabular}
    \end{small}
\end{table}

\begin{figure*}[ht] 
    \centerline{
    \begin{subfigure}[t]{0.51\columnwidth}
    \includegraphics[width=1\textwidth]{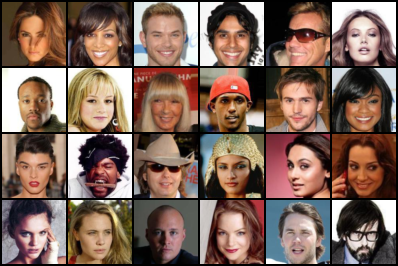}
    \caption{raw}
    \label{fig:celebA:samples:raw}
    \end{subfigure}
    \begin{subfigure}[t]{0.51\columnwidth}
    \includegraphics[width=1\textwidth]{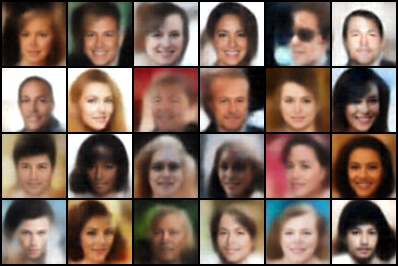}
    \caption{same samples gender privatized}
    \label{fig:celebA:samples:priv:gender}
    \end{subfigure}
    }
    \caption{Sampled images. We try to protect gender as the private attribute and keep the expression of smiling as the utility. We switch some female faces to males and also some males to females while preserving the celebrity's smile. The blurriness of the privatized images is due to the compactness of the latent representation that is generated from the VAE model and not from our privatization scheme. More details can be found in \figref{fig:celebA:raw:enc:dec:samples} in \secref{appx:sec:celebA:examples:more}.}
    \label{fig:celebA:gender:priv:examples}
\end{figure*}

\textbf{CelebA}: For the CelebA dataset, we consider the case when there exists many private and utility label combinations depending on the user's preferences. Specifically, we experiment on the private labels gender (male), age (young), attractive, eyeglasses, big nose, big lips, high cheekbones, or wavy hair, and we set the utility label as smiling for each private label to simplify the experiment. \tableref{tab:celebA:clf:acc:results} shows the classification results of multiple approaches. Our trained generative adversarial filter reduces the accuracy down to 73\% on average, which is only 6\% more than the worst case accuracy demonstrating the ability to protect the private attributes. Meanwhile, we only sacrifice a small amount of the utility accuracy (3\% drop), which ensures that the privatized data can still serve the desired classification tasks (All details are summarized in \tableref{tab:celebA:clf:acc:results}). We show samples of the gender-privatized images in \Figref{fig:celebA:gender:priv:examples}, which indicates the desired phenomenon that some female images are switched into male images and some male images are changed into female images. More example images on other privatized attributes can be found in appendix \secref{appx:sec:celebA:examples:more}.
%
%
\begin{table}[!bpht]
\caption{Classification accuracy on CelebA. The row \textbf{VAE-emb} is our initial classification accuracy on the latent vectors from our trained encoder. The row \textbf{Random-guess} demonstrates the worst possible classification accuracy. The row \textbf{VAE-g-filter} is the classification accuracy on the perturbed latent vectors yielded by our generative adversarial filters. The state-of-the-art classifier \citep{torfason2016face} can achieve 87\% accuracy on the listed private labels on average while our trained VAE can achieve a comparable accuracy (83\% on average) on the more compact latent representations. More importantly, our trained generative adversarial filter can reduce the accuracy down to 73\% on average, which is only 6\% more than the worst case accuracy demonstrating the ability to protect the private attributes. Meanwhile, we only sacrifice a small amount of the utility accuracy (3\% drop), which ensures that the privatized data can still serve the desired classification tasks}
\label{tab:celebA:clf:acc:results}
    \centerline{
    \begin{scriptsize}
    \begin{tabular}{c|cccccccc|c|c}
    \toprule
    & \multicolumn{9}{c|}{\textbf{Private attr.}} & \textbf{Utility attr.}\\
         & Male & Young & Attractive & H. Cheekbones & B. lips & B. nose & Eyeglasses & W. Hair & \textbf{Avg} & Smiling \\
    \midrule      
        \citet{liu2015faceattributes} & 0.98 & 0.87 & 0.81 & 0.87 & 0.68 & 0.78 & 0.99 & 0.80 &   0.84  &  0.92 \\ 
        \citet{torfason2016face} & 0.98 & 0.89 & 0.82 & 0.87 & 0.73 & 0.83 & 0.99 & 0.84 &  0.87 & 0.93 \\
        VAE-emb & 0.90 & 0.84 & 0.80 & 0.85 & 0.68 & 0.78 & 0.98 & 0.78 &    0.83 & 0.86 \\
        Random-guess & 0.51  &  0.76 & 0.54  &  0.54 & 0.67 & 0.77 & 0.93  &0.63 &   0.67 & 0.50 \\
        VAE-g-filter & 0.61 & 0.76 & 0.62 & 0.78 & 0.67 & 0.78 & 0.93 & 0.66 & 0.73 &   0.83 \\
        \bottomrule 
    \end{tabular}
    \end{scriptsize}
    }
\end{table}
\section{Discussion}
In order to clarify how our scheme can be run in a local and distributed fashion, we perform a basic experiment with 2 independent users to demonstrate this capability. The first user adopts the label of digit $\geq 5$ as private and odd or even as the utility label. The second user prefers the opposite and wishes to privatize odd or even and maintain $\geq 5$ as the utility label. We first partition the MNIST dataset into 10 equal parts where the first part belongs to one user and the second part belongs to the other user. The final eight parts have already been made suitable for public use either through privatization or because it does not contain information that their original owners have considered sensitive. It is then encoded into their 10 dimensional representations and passed onto the two users for the purpose of training an appropriate classifier rather than one trained on a single user's biased dataset. Since the data is already encoded into its representation, the footprint is very small when training the classifiers. Then, the generative filter for each user is trained separately and only on the single partition of personal data. Meanwhile, the adversarial and utility classifiers for each user are trained separately and only on the 8 parts of public data combined with the one part of personal data. The final result is 2 generative filters, one for each user, which corresponds to their own choice of private and utility labels. After the generative filters privatize their data, we can evaluate the classification accuracy on the private and utility labels as measured by adversaries trained on the full privatized dataset. \tableref{tab:mnist:distr} demonstrates the classification accuracy on the two users privatized data. This shows how multiple generative filters can be trained independently to successfully privatize small subsets of data.

\begin{table}[ht]
    \centering
    \caption{Accuracy of adversarial classifiers on two users private labels}
    \label{tab:mnist:distr}
    \begin{small}
    \begin{tabular}{c|c|c}
    \toprule 
    \textbf{Classifier type} & \textbf{User 1 (privatize $\mathbf{\geq 5}$)} & \textbf{User 2 (privatize odd)} \\
    \midrule 
        Private attr. & 0.679 & 0.651 \\
        Utility attr. & 0.896 & 0.855 \\
    \bottomrule    
    \end{tabular}
    \end{small}
\end{table}

We notice that our classification results on utility label (smiling) in the CelebA experiment performs worse than the state of the art classifiers presented in \cite{liu2015faceattributes} and \cite{torfason2016face}. However, the main purpose of our approach is not building the best image classifier. Instead of constructing a large feature vector (through convolution , pooling, and non-linear activation operations), we compress a face image down to a 50-dimension vector as the embedding\footnote{We use a VAE type architecture to compress the image down to a 100 dimensional vector, then enforce the first 50 dimensions as the mean and the second 50 dimensions as the variance}. We make the perturbation through a generative filter to yield a vector with the same dimensions. Finally, we construct a neural network with two fully-connected layers and an elu activation after the first layer to perform the classification task. We believe the deficit of the accuracy is due to the compact dimensionality of the representations and the simplified structure of the classifiers. We expect that a more powerful state of the art classifier trained on the released private images will still demonstrate the decreased accuracy on the private labels compared to the original non private images while maintaining higher accuracy on the utility labels. \\

We also discover an interesting connection between our idea of decorrelating the released data with sensitive attributes and the notion of learning fair representations\citep{pmlr-v28-zemel13, louizos2015variational,madras2018learning}. In fairness learning, the notion of demographic parity requires the prediction of a target output to be independent with respect to certain protected attributes. We find our generative filter could be used to produce a fair representation after privatizing the raw embedding, which shares the similar idea of demographic parity. Other notions in fairness learning such as equal odds and equal opportunity \citep{hardt2016equality} will be left for future work.        

%
\section{Conclusion}
In this paper, we propose an architecture for privatizing data while maintaining the utility that decouples for use in a distributed manner. Rather than training a very deep neural network or imposing a particular discriminator to judge real or fake images, we rely on a pretrained VAE that can create a comprehensive low dimensional representation from the raw data. We then find smart perturbations in the latent space according to customized requirements (e.g. various choices of the private label and utility label), using a robust optimization approach. Such an architecture and procedure enables small devices such as mobile phones or home assistants (e.g. Google home mini) to run a light weight learning algorithm to privatize data under various settings on the fly.

%


\bibliography{SafeML_ICLRworkshop_Ref}

\begin{thebibliography}{39}
\providecommand{\natexlab}[1]{#1}
\providecommand{\url}[1]{\texttt{#1}}
\expandafter\ifx\csname urlstyle\endcsname\relax
  \providecommand{\doi}[1]{doi: #1}\else
  \providecommand{\doi}{doi: \begingroup \urlstyle{rm}\Url}\fi

\bibitem[Abadi \& Andersen(2016)Abadi and Andersen]{abadi2016learning}
Mart{\'\i}n Abadi and David~G Andersen.
\newblock Learning to protect communications with adversarial neural
  cryptography.
\newblock \emph{arXiv preprint arXiv:1610.06918}, 2016.

\bibitem[Beutel et~al.(2017)Beutel, Chen, Zhao, and Chi]{beutel2017data}
Alex Beutel, Jilin Chen, Zhe Zhao, and Ed~H Chi.
\newblock Data decisions and theoretical implications when adversarially
  learning fair representations.
\newblock \emph{arXiv preprint arXiv:1707.00075}, 2017.

\bibitem[Chen et~al.(2018)Chen, Kairouz, and Rajagopal]{chen2018understanding}
Xiao Chen, Peter Kairouz, and Ram Rajagopal.
\newblock Understanding compressive adversarial privacy.
\newblock In \emph{2018 IEEE Conference on Decision and Control (CDC)}, pp.\
  6824--6831. IEEE, 2018.

\bibitem[Cichocki \& Amari(2010)Cichocki and Amari]{cichocki2010families}
Andrzej Cichocki and Shun-ichi Amari.
\newblock Families of alpha-beta-and gamma-divergences: Flexible and robust
  measures of similarities.
\newblock \emph{Entropy}, 12\penalty0 (6):\penalty0 1532--1568, 2010.

\bibitem[De~Boer et~al.(2005)De~Boer, Kroese, Mannor, and
  Rubinstein]{de2005tutorial}
Pieter-Tjerk De~Boer, Dirk~P Kroese, Shie Mannor, and Reuven~Y Rubinstein.
\newblock A tutorial on the cross-entropy method.
\newblock \emph{Annals of operations research}, 134\penalty0 (1):\penalty0
  19--67, 2005.

\bibitem[Dua \& Graff(2017)Dua and Graff]{Dua:2019}
Dheeru Dua and Casey Graff.
\newblock {UCI} machine learning repository, 2017.
\newblock URL \url{http://archive.ics.uci.edu/ml}.

\bibitem[Dupont(2018)]{dupont2018learning}
Emilien Dupont.
\newblock Learning disentangled joint continuous and discrete representations.
\newblock In \emph{Advances in Neural Information Processing Systems}, pp.\
  708--718, 2018.

\bibitem[Dwork(2011)]{dwork2011differential}
Cynthia Dwork.
\newblock Differential privacy.
\newblock \emph{Encyclopedia of Cryptography and Security}, pp.\  338--340,
  2011.

\bibitem[Dwork et~al.(2006)Dwork, McSherry, Nissim, and
  Smith]{dwork2006calibrating}
Cynthia Dwork, Frank McSherry, Kobbi Nissim, and Adam Smith.
\newblock Calibrating noise to sensitivity in private data analysis.
\newblock In \emph{Theory of cryptography conference}, pp.\  265--284.
  Springer, 2006.

\bibitem[Dwork et~al.(2014)Dwork, Roth, et~al.]{dwork2014algorithmic}
Cynthia Dwork, Aaron Roth, et~al.
\newblock The algorithmic foundations of differential privacy.
\newblock \emph{Foundations and Trends{\textregistered} in Theoretical Computer
  Science}, 9\penalty0 (3--4):\penalty0 211--407, 2014.

\bibitem[Edwards \& Storkey(2015)Edwards and Storkey]{edwards2015censoring}
Harrison Edwards and Amos Storkey.
\newblock Censoring representations with an adversary.
\newblock \emph{arXiv preprint arXiv:1511.05897}, 2015.

\bibitem[Gao et~al.(2015)Gao, Ver~Steeg, and Galstyan]{gao2015efficient}
Shuyang Gao, Greg Ver~Steeg, and Aram Galstyan.
\newblock Efficient estimation of mutual information for strongly dependent
  variables.
\newblock In \emph{Artificial Intelligence and Statistics}, pp.\  277--286,
  2015.

\bibitem[Goodfellow et~al.(2014)Goodfellow, Pouget-Abadie, Mirza, Xu,
  Warde-Farley, Ozair, Courville, and Bengio]{goodfellow2014generative}
Ian Goodfellow, Jean Pouget-Abadie, Mehdi Mirza, Bing Xu, David Warde-Farley,
  Sherjil Ozair, Aaron Courville, and Yoshua Bengio.
\newblock Generative adversarial nets.
\newblock In \emph{Advances in neural information processing systems}, pp.\
  2672--2680, 2014.

\bibitem[Hamm(2017)]{hamm2017minimax}
Jihun Hamm.
\newblock Minimax filter: learning to preserve privacy from inference attacks.
\newblock \emph{The Journal of Machine Learning Research}, 18\penalty0
  (1):\penalty0 4704--4734, 2017.

\bibitem[Hardt et~al.(2016)Hardt, Price, Srebro, et~al.]{hardt2016equality}
Moritz Hardt, Eric Price, Nati Srebro, et~al.
\newblock Equality of opportunity in supervised learning.
\newblock In \emph{Advances in neural information processing systems}, pp.\
  3315--3323, 2016.

\bibitem[Higgins et~al.(2017)Higgins, Matthey, Pal, Burgess, Glorot, Botvinick,
  Mohamed, and Lerchner]{higgins2017beta}
Irina Higgins, Loic Matthey, Arka Pal, Christopher Burgess, Xavier Glorot,
  Matthew Botvinick, Shakir Mohamed, and Alexander Lerchner.
\newblock beta-vae: Learning basic visual concepts with a constrained
  variational framework.
\newblock In \emph{International Conference on Learning Representations}, 2017.

\bibitem[Huang et~al.(2017{\natexlab{a}})Huang, Kairouz, Chen, Sankar, and
  Rajagopal]{huang2017context}
Chong Huang, Peter Kairouz, Xiao Chen, Lalitha Sankar, and Ram Rajagopal.
\newblock Context-aware generative adversarial privacy.
\newblock \emph{Entropy}, 19\penalty0 (12):\penalty0 656, 2017{\natexlab{a}}.

\bibitem[Huang et~al.(2017{\natexlab{b}})Huang, Liu, Van Der~Maaten, and
  Weinberger]{huang2017densely}
Gao Huang, Zhuang Liu, Laurens Van Der~Maaten, and Kilian~Q Weinberger.
\newblock Densely connected convolutional networks.
\newblock In \emph{CVPR}, volume~1, pp.\ ~3, 2017{\natexlab{b}}.

\bibitem[Kingma \& Ba(2014)Kingma and Ba]{kingma2014adam}
Diederik~P Kingma and Jimmy Ba.
\newblock Adam: A method for stochastic optimization.
\newblock \emph{arXiv preprint arXiv:1412.6980}, 2014.

\bibitem[Kingma \& Welling(2013)Kingma and Welling]{kingma2013auto}
Diederik~P Kingma and Max Welling.
\newblock Auto-encoding variational bayes.
\newblock \emph{arXiv preprint arXiv:1312.6114}, 2013.

\bibitem[Koh et~al.(2018)Koh, Steinhardt, and Liang]{koh2018stronger}
Pang~Wei Koh, Jacob Steinhardt, and Percy Liang.
\newblock Stronger data poisoning attacks break data sanitization defenses.
\newblock \emph{arXiv preprint arXiv:1811.00741}, 2018.

\bibitem[Krizhevsky et~al.(2012)Krizhevsky, Sutskever, and
  Hinton]{krizhevsky2012imagenet}
Alex Krizhevsky, Ilya Sutskever, and Geoffrey~E Hinton.
\newblock Imagenet classification with deep convolutional neural networks.
\newblock In \emph{Advances in neural information processing systems}, pp.\
  1097--1105, 2012.

\bibitem[LeCun \& Cortes(2010)LeCun and
  Cortes]{lecun-mnisthandwrittendigit-2010}
Yann LeCun and Corinna Cortes.
\newblock {MNIST} handwritten digit database.
\newblock 2010.
\newblock URL \url{http://yann.lecun.com/exdb/mnist/}.

\bibitem[Liu et~al.(2017)Liu, Chakraborty, and Mittal]{liu2017deeprotect}
Changchang Liu, Supriyo Chakraborty, and Prateek Mittal.
\newblock Deeprotect: Enabling inference-based access control on mobile sensing
  applications.
\newblock \emph{arXiv preprint arXiv:1702.06159}, 2017.

\bibitem[Liu et~al.(2015)Liu, Luo, Wang, and Tang]{liu2015faceattributes}
Ziwei Liu, Ping Luo, Xiaogang Wang, and Xiaoou Tang.
\newblock Deep learning face attributes in the wild.
\newblock In \emph{Proceedings of International Conference on Computer Vision
  (ICCV)}, 2015.

\bibitem[Louizos et~al.(2015)Louizos, Swersky, Li, Welling, and
  Zemel]{louizos2015variational}
Christos Louizos, Kevin Swersky, Yujia Li, Max Welling, and Richard Zemel.
\newblock The variational fair autoencoder.
\newblock \emph{arXiv preprint arXiv:1511.00830}, 2015.

\bibitem[Madras et~al.(2018)Madras, Creager, Pitassi, and
  Zemel]{madras2018learning}
David Madras, Elliot Creager, Toniann Pitassi, and Richard Zemel.
\newblock Learning adversarially fair and transferable representations.
\newblock \emph{arXiv preprint arXiv:1802.06309}, 2018.

\bibitem[Madry et~al.(2017)Madry, Makelov, Schmidt, Tsipras, and
  Vladu]{madry2017towards}
Aleksander Madry, Aleksandar Makelov, Ludwig Schmidt, Dimitris Tsipras, and
  Adrian Vladu.
\newblock Towards deep learning models resistant to adversarial attacks.
\newblock \emph{arXiv preprint arXiv:1706.06083}, 2017.

\bibitem[McInnes et~al.(2018)McInnes, Healy, and Melville]{mcinnes2018umap}
Leland McInnes, John Healy, and James Melville.
\newblock Umap: Uniform manifold approximation and projection for dimension
  reduction.
\newblock \emph{arXiv preprint arXiv:1802.03426}, 2018.

\bibitem[McMahan et~al.(2016)McMahan, Moore, Ramage, Hampson,
  et~al.]{mcmahan2016communication}
H~Brendan McMahan, Eider Moore, Daniel Ramage, Seth Hampson, et~al.
\newblock Communication-efficient learning of deep networks from decentralized
  data.
\newblock \emph{arXiv preprint arXiv:1602.05629}, 2016.

\bibitem[Nguyen et~al.(2010)Nguyen, Wainwright, and
  Jordan]{nguyen2010estimating}
XuanLong Nguyen, Martin~J Wainwright, and Michael~I Jordan.
\newblock Estimating divergence functionals and the likelihood ratio by convex
  risk minimization.
\newblock \emph{IEEE Transactions on Information Theory}, 56\penalty0
  (11):\penalty0 5847--5861, 2010.

\bibitem[Nowozin et~al.(2016)Nowozin, Cseke, and Tomioka]{nowozin2016f}
Sebastian Nowozin, Botond Cseke, and Ryota Tomioka.
\newblock f-gan: Training generative neural samplers using variational
  divergence minimization.
\newblock In \emph{Advances in neural information processing systems}, pp.\
  271--279, 2016.

\bibitem[Oshri et~al.(2018)Oshri, Hu, Adelson, Chen, Dupas, Weinstein, Burke,
  Lobell, and Ermon]{oshri2018infrastructure}
Barak Oshri, Annie Hu, Peter Adelson, Xiao Chen, Pascaline Dupas, Jeremy
  Weinstein, Marshall Burke, David Lobell, and Stefano Ermon.
\newblock Infrastructure quality assessment in africa using satellite imagery
  and deep learning.
\newblock In \emph{Proceedings of the 24th ACM SIGKDD International Conference
  on Knowledge Discovery \& Data Mining}, pp.\  616--625. ACM, 2018.

\bibitem[Papernot et~al.(2017)Papernot, McDaniel, Goodfellow, Jha, Celik, and
  Swami]{papernot2017practical}
Nicolas Papernot, Patrick McDaniel, Ian Goodfellow, Somesh Jha, Z~Berkay Celik,
  and Ananthram Swami.
\newblock Practical black-box attacks against machine learning.
\newblock In \emph{Proceedings of the 2017 ACM on Asia Conference on Computer
  and Communications Security}, pp.\  506--519. ACM, 2017.

\bibitem[Rezaei et~al.(2018)Rezaei, Xiao, Gao, and Li]{rezaei2018protecting}
Aria Rezaei, Chaowei Xiao, Jie Gao, and Bo~Li.
\newblock Protecting sensitive attributes via generative adversarial networks.
\newblock \emph{arXiv preprint arXiv:1812.10193}, 2018.

\bibitem[Song et~al.(2018)Song, Kalluri, Grover, Zhao, and
  Ermon]{song2018learning}
Jiaming Song, Pratyusha Kalluri, Aditya Grover, Shengjia Zhao, and Stefano
  Ermon.
\newblock Learning controllable fair representations.
\newblock \emph{arXiv preprint arXiv:1812.04218}, 2018.

\bibitem[Torfason et~al.(2016)Torfason, Agustsson, Rothe, and
  Timofte]{torfason2016face}
Robert Torfason, Eirikur Agustsson, Rasmus Rothe, and Radu Timofte.
\newblock From face images and attributes to attributes.
\newblock In \emph{Asian Conference on Computer Vision}, pp.\  313--329.
  Springer, 2016.

\bibitem[Wong \& Kolter(2018)Wong and Kolter]{wong2018provable}
Eric Wong and Zico Kolter.
\newblock Provable defenses against adversarial examples via the convex outer
  adversarial polytope.
\newblock In \emph{International Conference on Machine Learning}, pp.\
  5283--5292, 2018.

\bibitem[Zemel et~al.(2013)Zemel, Wu, Swersky, Pitassi, and
  Dwork]{pmlr-v28-zemel13}
Richard Zemel, Yu~Wu, Kevin Swersky, Toni Pitassi, and Cynthia Dwork.
\newblock Learning fair representations.
\newblock In Sanjoy Dasgupta and David McAllester (eds.), \emph{Proceedings of
  the 30th International Conference on Machine Learning}, volume~28 of
  \emph{Proceedings of Machine Learning Research}, pp.\  325--333, Atlanta,
  Georgia, USA, 17--19 Jun 2013. PMLR.
\newblock URL \url{http://proceedings.mlr.press/v28/zemel13.html}.

\end{thebibliography}
\bibliographystyle{iclr2019_conference}

\clearpage
\begin{center}
{\Large Supplement Material}
\end{center}
\section{Appendix}

\subsection{VAE training}
\label{supp:tech:approch:vae:explained}
A variational autoencoder is a generative model defining a joint probability distribution between a latent variable ${z}$ and original input ${x}$. We assume the data is generated from a parametric distribution $p({x}|{z} ; \theta)$ that depends on latent variable ${z}$, where $\theta$ are the parameters of a neural network, which usually is a decoder net. Maximizing the marginal likelihood $p({x}|{z}; \theta)$ directly is usually intractable. Thus, we use the variational inference method proposed by \cite{kingma2013auto} to optimize $\log p({x}|{z}; \theta)$ over an alternative distribution $q({z}|{x}; \phi)$ with an additional KL divergence term $D_{KL}(q({z}|{x}; \phi) || p({z})\big)$ , where the $\phi$ are parameters of a neural net and $p({z})$ is an assumed prior over the latent space. The resulting cost function is often called evidence lower bound (ELBO)
\begin{align*}
    \log p({x}; \theta) & \geq  \mathbb{E}_{q({z}|{x};\phi)} [\log p({x}|{z};\theta)] - D_{KL}\big(q({z}|{x};\phi) || p({z})\big) = \mathcal{L}_{\text{ELBO}} .
\end{align*}
Maximizing the ELBO is implicitly maximizing the log-likelihood of $p(\mathbf{x})$. The negative objective (also known as negative ELBO) can be interpreted as minimizing the reconstruction loss of a probabilistic autoencoder and regularizing the posterior distribution towards a prior.
Although the loss of the VAE is mentioned in many studies \citep{kingma2013auto, louizos2015variational}, we include the derivation of the following relationships for completeness of the context
\begin{align*}
    D_{KL}(q(z)||p(z|x; \theta)) = \sum_{z} q(z) \log \frac{q(z)}{p(z|x;\theta)} = -\sum_{z}
q(z) \log p(z, x;\theta) + \log p(x; \theta)  \underbrace{- H(q)}_{\sum_{z} q(z) \log q(z)} \geq 0
\end{align*}

The Evidence lower bound (ELBO) for any distribution $q$ 

\begin{align*}
    & \log p(x; \theta) \geq \sum_{z}
q(z) \log p(z, x;\theta) -  \sum_{z} q(z) \log q(z)  \quad [\text{since } D_{KL}(\cdot, \cdot) \geq 0] \\
& = \sum_{z}
q(z)\Big( \log p(z, x;\theta) - \log p(z) \Big)  - \sum_{z} q(z)\Big(   \log q(z) - \log p(z)\Big) \\
& \explainEq{i} \mathbb{E}_{q(z|x; \phi)}[\log p(x|z; \theta)] - D_{KL}\big(q(z|x; \phi)||p(z)\big) = \mathcal{L}_\text{ELBO}, 
\end{align*}
where (i) holds because we treat encoder net $q(z|x; \phi)$ as the distribution $q(z)$. By placing the corresponding parameters of the encoder and decoder networks, and the negative sign on the ELBO expression, we get the loss function \eqrefp{eq:vae:nelbo}. The architecture of the encoder and decoder for the {MNIST} experiments is explained in \secref{supp:nn:archi:explain}. 
%

In our experiments with the {MNIST} dataset, the negative ELBO objective works well because each pixel value (0 black or 1 white) is generated from a Bernoulli distribution. However, in the experiments of {CelebA}, we change the reconstruction loss into the $\ell_p$ norm of the difference between raw and reconstructed samples because the RGB pixels are not Bernoulli random variables. We still add the regularization KL term as follows  
\begin{align*}
    \mathbb{E}_{x^{\prime} \sim p({x}|{z}; \theta)}\big[||x - x^{\prime}||_{\text{p}}\big] + \gamma D_{\text{KL}}\big(q({z}|{x};\phi) || p({z})\big). 
\end{align*}
Throughout the experiments we use a Gaussian $\mathcal{N}(0, \mathbf{I})$ as the prior $p({z})$, $x$ is sampled from the data, and $\gamma$ is a hyper-parameter. The reconstruction loss uses the $\ell_2$ norm by default, although, the $\ell_1$ norm is acceptable too.

When training the VAE, we additionally ensure that small perturbations in the latent space will not yield huge deviations in the reconstructed space. More specifically, we denote the encoder and decoder to be $g_e$ and $g_d$ respectively. The generator $g$ can be considered as a composition of an encoding and decoding process, i.e. $g(X)$ = $(g_d \circ g_e) (X) = g_d(g_e(X))$, where we ignore the $Y$ inputs here for the purpose of simplifying the explanation. One desired intuitive property for the decoder is to maintain that small changes in the input latent space $\mathcal{Z}$ still produce plausible faces similar to the original latent space when reconstructed. Thus, we would like to impose some Lipschitz continuity property on the decoder, i.e. for two points $z^{(1)}, z^{(2)} \in \mathcal{Z}$, we assume $|| g_d(z^{(1)}) - g_d(z^{(2)}) || \leq C_L ||z^{(1)} - z^{(2)} ||$ where $C_L$ is some Lipschitz constant (or equivalently $||\nabla_z g_d(z)|| \leq C_L$). In the implementation of our experiments, the gradient for each batch (with size $m$) is 
\begin{align*}
    \nabla_{z} g_d(z) = \begin{bmatrix} 
        \frac{\partial g_d(z^{(1)})}{\partial z^{(1)}} \\
        \vdots \\
         \frac{\partial g_d(z^{(m)})}{\partial z^{(m)}}
    \end{bmatrix} . 
\end{align*}
It is worth noticing that $ 
   \frac{\partial g_d(z^{(i)})}{\partial z^{(i)}} =  \frac{\partial \sum_{i=1}^{m}g_d(z^{(i)})}{\partial z^{(i)}}$,   
because  $\frac{\partial g_d(z^{(j)})}{\partial z^{(i)}}$ = 0 when $i\neq j$. Thus, we define $Z$ as the batched latent input, 
$
    Z = \begin{bmatrix}
    z^{(1)} \\
    \vdots \\
    z^{(m)}
    \end{bmatrix}$,
and use $
    \nabla_{z} g_d(z) = \frac{\partial}{\partial Z} \sum_{i=1}^{m} g_d(z^{(i)}), 
$ to avoid the iterative calculation of each gradient within a batch. The loss used for training the VAE is modified to be 
\begin{small}
\begin{align}
    \mathbb{E}_{x^{\prime} \sim p({x}|{z}; \theta_d)}\big[||x - x^{\prime}||_{2}\big] + \gamma D_{\text{KL}}\big(q({z}|{x};\theta_e) || p({z})\big) + \kappa  \mathbb{E}_{z \sim r_{\alpha}(z)}\big[ \big(||\nabla_z (g_{\theta_d}(z)||_2 - C_L\big)_{+}^2 \big],  \label{eq:supp:vae:loss:variant}
\end{align}
\end{small}
where $x$ are samples drawn from the image data, $\gamma, \kappa$, and $C_L$ are hyper-parameters, and $(x)_{+}$ means $\max\{x, 0\}$. Finally, $r_\alpha$ is defined by sampling $\alpha \sim \textsf{Unif}[0,1]$,  $Z_{1} \sim p(z)$, and  $Z_{2} \sim q(z|x)$, and returning $\alpha Z_{1} + (1-\alpha) Z_{2}$. We optimize over $r_\alpha$ to ensure that points in between the prior and learned latent distribution maintain a similar reconstruction to points within the learned distribution. This gives us the Lipschitz continuity properties we desire for points perturbed outside of the original learned latent distribution.

\subsection{Robust Optimization \& adversarial training} \label{supp:tech:approach:noise_inject:train}
In this part, we formulate the generator training as a robust optimization. Essentially, the generator is trying to construct a new latent distribution that reduces the correlation between data samples and sensitive labels while maintaining the correlation with utility labels by leveraging the appropriate generative filters. The new latent distribution, however, cannot deviate from the original distribution too much (bounded by $b$) to maintain the general quality of the reconstructed images. To simplify the notation, we will use $h$ for the classifier $h_{\theta_h}$ (similar notion applies to $\nu$ or $\nu_{\theta_{\nu}}$). We also consider the triplet $(\tilde{z}, y, u)$ as the input data, where $\tilde{z}$ is the perturbed version of the original embedding $z$, which is the latent representation of image $x$. The values $y$ and $u$ are the sensitive label and utility label respectively. Without loss of generality, we succinctly express the loss $\ell(h; (\tilde{z}, y))$ as $\ell(h; \tilde{z})$ (similarly expressing $\tilde{\ell}(\nu; (\tilde{z}, u))$ as $\tilde{\ell}(\nu; \tilde{z})$). We assume the sample input $\tilde{z}$ follows the distribution $q_{\psi}$ that needs to be determined. Thus, the robust optimization is   
\begin{align}
    & \max_{q_{\psi}}\big\{\min_{h} \mathbb{E}_{q_{\psi}}\big(\ell(h; \tilde{z} ) \big) - \beta\min_{\nu} \mathbb{E}_{q_{\psi}} \big( \tilde{\ell}(\nu; \tilde{z} ) \big) \big\} \\ 
    s.t & \qquad D_f(q_{\psi}||q_{\phi} ) \leq b, \\
    & \qquad \mathbb{E}_{q_{\phi}} = 1,
\end{align}
where $q_{\phi}$ is the distribution of the raw embedding $z$ (also known as $q_{\theta_e}(z|x)$). The $f$-divergence \citep{nguyen2010estimating, cichocki2010families} between $q_{\psi}$ and $q_{\phi}$ is $D_f(q_{\psi}||q_{\phi}) = \int q_\phi(z)f(\frac{q_{\psi}(z)}{q_{\phi}(z)}) d\mu(z) $ (assuming $q_{\psi}$ and $q_{\phi}$ are absolutely continuous with respect to measure $\mu$). A few typical divergences \citep{nowozin2016f}, depending on the choices of $f$, are
\begin{enumerate}
    \item KL-divergence $D_{\text{KL}}(q_{\psi}||q_{\phi}) \leq b$, by taking $f(t)=t\log t$
    \item reverse KL-divergence $D_{KL}(q_{\phi}||q_{\psi})\leq  b$, by taking $f(t) =-\log t$
    \item $\chi^2$-divergence $D_{\chi^2}(q_{\psi}||q_{\phi}) \leq  b$, by taking $ f(t)=\frac{1}{2}(t-1)^2$. 
\end{enumerate}
In the remainder of this section, we focus on the KL and $\chi^2$ divergence to build a connection between the divergence based constraint we use and some norm-ball based constraint seen in \citet{wong2018provable,madry2017towards, koh2018stronger}, and \citet{papernot2017practical}.
\subsubsection{Extension to multivariate Gaussian}
When we train the VAE in the beginning, we impose the distribution $q_{\phi}$ to be a multivariate Gaussian by penalizing the KL-divergence between $q_{\theta_e}(z|x)$ and a prior normal distribution $\mathcal{N}(0, \mathbf{I})$, where $\mathbf{I}$ is the identity matrix. Without loss of generality, we can assume the raw encoded variable $z$ follows a distribution $q_{\phi}$ that is the Gaussian distribution $\mathcal{N}(\mu_1, \Sigma_1)$ (more precisely $\mathcal{N}\big(\mu_1(x), \Sigma_1(x)\big)$, where the mean and variance depends on samples $x$, but we suppress the $x$ to simplify notation). The new perturbed distribution $q_{\psi}$ is then also a Gaussian $\mathcal{N}(\mu_2, \Sigma_2)$. Thus, the constraint for the KL divergence becomes 
\begin{align*}
    & D_{\text{KL}}(q_{\psi}||q_{\phi}) = \mathbb{E}_{q_{\psi}}(\log \frac{q_{\psi}}{q_{\phi}}) \\
    & = \frac{1}{2}\mathbb{E}_{q_{\psi}}[-\log \det(\Sigma_2) - (z - \mu_2)^T\Sigma_2^{-1}(z-\mu_2) + \log \det(\Sigma_1) + (z-\mu_1)^T\Sigma_1^{-1}(z-\mu_1) ] \\
    & = \frac{1}{2}\big[ \log\frac{\det (\Sigma_1)}{\det (\Sigma_2)} - \underbrace{d}_{=\Tr(\mathbf{I})} + \Tr(\Sigma_1^{-1}\Sigma_2) + (\mu_1 - \mu_2 )^T\Sigma_1^{-1}(\mu_1 - \mu_2) \big] \leq b.
\end{align*}
If we further consider the case that $\Sigma_2 = \Sigma_1 = \Sigma$, then
\begin{align}
    D_{KL}(q_{\psi}||q_{\phi}) = \frac{1}{2}(\mu_1 - \mu_2)^T \Sigma^{-1} (\mu_1 - \mu_2) = \frac{1}{2} ||\mu_1 - \mu_2||_{\Sigma^{-1}}^2 \leq b \label{supp:eq:D_KL:two:gaussian}.
\end{align}
When $\Sigma=\mathbf{I}$, the preceding constraint is equivalent to $\frac{1}{2}||\mu_1 - \mu_2 ||_2^2$. It is worth mentioning that such a divergence based constraint is also connected to the norm-ball based constraint on samples. When the variance of $\tilde{z}^Tz$ is not larger than the total variance of $z^Tz$, denoted as $\Var(\tilde{z}^Tz) = \zeta \leq \Tr(\Sigma) = \Var(z^Tz)$, it can be discovered that
\begin{align*}
    ||\mu_1 - \mu_2 ||_2^2  
    \explainLeq{i}  
    \mu_1^T\mu_1 + \mu_2^T\mu_2 - 2\mu_1^T\mu_2 + 2\Tr(\Sigma) - 2\zeta
    =
    \mathbb{E}_{z\sim q_{\phi}, \tilde{z}\sim q_{\psi}}\big(||z - \tilde{z}||_2^2\big) , 
\end{align*}
where (i) uses the assumption $\Tr(\Sigma) - \zeta \geq 0 $. This implies that the norm-ball ($l_2$-norm) based perturbation constraint $\frac{1}{2}\mathbb{E}\big(||z-\tilde{z}||_2^2\big) \leq b $ automatically satisfies $D_{KL}(q_{\psi}||q_{\phi}) \leq b $. In the other case where $\Tr(\Sigma) - \zeta \leq 0$; constraint satisfaction is switched, so the $D_{KL}(q_{\psi}||q_{\phi}) \leq b $ constraint automatically satisfies  $\frac{1}{2}\mathbb{E}\big(||z-\tilde{z}||_2^2\big) \leq b $.

In the case of $\chi^2$-divergence,

\begin{align*}
    & D_{\chi^2}(q_{\psi}||q_{\phi})  = E_{q_{\phi}}[\frac{1}{2}(\frac{q_{\psi}}{q_{\phi}}-1)^2] \\
    & =  \frac{1}{2}\Bigg[ \frac{\det(\Sigma_1)}{\det(\Sigma_2)} \exp\big(-\Tr(\Sigma_2^{-1}\Sigma_1) + \underbrace{d}_{=\Tr(\mathbf{I})} - \mu_1^T\Sigma_1^{-1}\mu_1 - \mu_2^T\Sigma_2^{-1}\mu_2 + 2\mu_1^T\Sigma_2^{-1}\mu_2 \big) \\ 
    & - 2 \frac{\det(\Sigma_1)^{\frac{1}{2}}}{\det(\Sigma_2)^{\frac{1}{2}}} \exp\big(-\frac{1}{2}\Tr(\Sigma_2^{-1}\Sigma_1) + \frac{d}{2}  -\frac{1}{2}\mu_1^T\Sigma_1^{-1}\mu_1 - \frac{1}{2}\mu_2^T\Sigma_2^{-1}\mu_2 + \mu_1^T\Sigma_2^{-1}\mu_2 \big) + 1\Bigg] . 
\end{align*}
When $\Sigma_1 = \Sigma_2 = \Sigma$, we have the following simplified expression 
\begin{align*}
    D_{\chi^2}(q_{\psi}||q_{\phi}) & =  \frac{1}{2}\Big[\exp\big(-||\mu_1 - \mu_2||_{\Sigma^{-1}}^2\big) -2 \exp\big(-\frac{1}{2}||\mu_1 - \mu_2||_{\Sigma^{-1}}^2\big)+ 1\Big] \\
    & = \frac{1}{2}[e^{-2s} - 2 e^{-s} + 1 ] = \frac{1}{2}(e^{-s}-1)^2 
\end{align*}
where $s = \frac{1}{2}||\mu_1 -\mu_2||_{\Sigma^{-1}}^2$. Letting $\frac{1}{2}(e^{-s}-1)^2 \leq b$ indicates $ -\sqrt{2b} \leq (e^{-s}-1) \leq \sqrt{2b}$. Since the value of $s$ is always non negative as a norm, $s\geq 0 \implies e^{-s} - 1 \leq 0 $. Thus, we have $ s \leq -\log\big((1-\sqrt{2b})_{+}\big)  $. Therefore, when the divergence constraint $D_{\chi^2}(q_{\psi}||q_{\phi}) \leq b$ is satisfied, we will have $ ||\mu_1 -\mu_2||_{\Sigma^{-1}}^2 \leq -2\log\big((1-\sqrt{2b})_{+}\big)$, which is similar to \eqrefp{supp:eq:D_KL:two:gaussian} with a new constant for the distortion budget.

We make use of these relationships in our implementation as follows. We define $\mu_1$ and $\mu_2$ to be functions that split the last layer of the output of the encoder part of the pretrained VAE, $g_{\theta_e}(x)$, and take the first half portion as the mean of the latent distribution. We let $\sigma_1$ be a function that takes the second half portion to be the (diagonal) of the variance of the latent distribution. Then, our implementation of  $||\mu_1\big(g_{\theta_e}(x)\big) - \mu_2\big(g_{\theta_f} (g_{\theta_e}(x), \epsilon, y)\big)  ||_{\sigma_1(g_{\theta_e}(x)) ^{-1}}^2 \leq b$ is equivalent to $ ||\mu_1 -\mu_2||_{\Sigma^{-1}}^2 \leq b $. As shown in the previous two derivations, optimizing over this constraint is similar to optimizing over the defined KL and $\chi^2$-divergence constraints.
 
%
\subsection{Comparison with Differential Privacy}
\label{supp:tech:approach:diff:privacy}
The basic intuition behind (differential) privacy is that it is very hard to tell whether the released sample $\tilde{z}$ originated from raw sample $x$ or $x'$, thus, protecting the privacy of the raw samples. Designing such a releasing scheme, which is also often called a channel (or mechanism), usually requires some randomized response or noise perturbation. The goal of such a releasing scheme does not necessarily involve reducing the correlation between released data and sensitive labels. Yet, it is worth investigating the notion of differential privacy and comparing the empirical performance of a typical scheme to our setting due to its prevalence in traditional privacy literature. In this section, we use the words channel, scheme, mechanism, and filter interchangeably as they have the same meaning in our context. Also, we overload the notation of $\varepsilon$ and $\delta$ because of the conventions in differential privacy literature. 

\begin{definition}{$\big( (\varepsilon, \delta)$-differential privacy \citep{dwork2014algorithmic} $\big)$ }
Let $\varepsilon, \delta \geq 0$. A channel $Q$ from space $\mathcal{X}$ to output space $\mathcal{Z}$ is differentially private if for all measurable sets $S \subset \mathcal{Z} $ and all neighboring samples $\{x_1, \hdots, x_n \}=x_{1:n} \in \mathcal{X} $ and $\{x_1', \hdots, x_n' \}=x_{1:n}' \in \mathcal{X} $, 
\begin{align}
    Q(Z\in S | x_{1:n}) \leq e^{\varepsilon} Q(Z\in S | x_{1:n}') + \delta .
\end{align}
\end{definition}
An alternative way to interpret this definition is that with high probability $1-\delta$, we have the bounded likelihood ratio $e^{-\varepsilon}\leq \frac{Q(z|x_{1:n})}{Q(z|x_{1:n}')} \leq e^{\varepsilon} $ (The likelihood ratio is close to 1 as $\varepsilon$ goes to $0$)\footnote{ Alternatively, we may express it as the probability $ P (|\log \frac{Q(z|x_{1:n})}{Q(z|x_{1:n}')}| > \varepsilon) \leq \delta$}. Consequently, it is difficult to tell if the observation $z$ is from $x$ or $x'$ if the ratio is close to 1. In the ongoing discussion, we consider the classical Gaussian mechanism $\tilde{z} = f(z) + w$,  where $f$ is some function (or query) that is defined on the latent space and $w \sim \mathcal{N}(0, \sigma^2\mathbf{I})$.  We first include a theorem from \citet{dwork2014algorithmic} to disclose how differential privacy using the Gaussian mechanism can be satisfied by our baseline implementation and constraints in the robust optimization formulation. We denote a pair of neighboring inputs as $z\simeq z'$ for abbreviation. 

\begin{theorem}{$\big($\cite{dwork2014algorithmic} Theorem A.1$\big)$}
\label{supp:thm:gaussian:mech:dp}
For any $\varepsilon, \delta \in (0, 1)$, the Gaussian mechanism with parameter $\sigma \geq \frac{L \sqrt{2\log(\frac{1.25}{\delta})}}{\varepsilon}$ is $(\varepsilon, \delta)$-differential private, where $L = \max_{z \simeq z'}||f(z) - f(z')||_2$ denotes the $l_2$-sensitivity of $f$. 
\end{theorem}

In our baseline implementation of the Gaussian mechanism, we apply the additive Gaussian noise on the samples $z$ directly to yield 
\begin{align}
    \tilde{z}=z+w
\end{align} (i.e. $f$ is the identity mapping). If $z \sim \mathcal{N}(\mu_1, \Sigma_1)$, we have $\tilde{z} \sim \mathcal{N}(\mu_1, \Sigma_1 + \sigma^2\mathbf{I})$. Therefore, considering KL-divergence as an example, we have  
\begin{align}
    D_{\text{KL}}(q_{\psi}||q_{\phi}) & = D_{\text{KL}}\big(\mathcal{N}(\mu_1, \Sigma_1 + \sigma^2\mathbf{I} )||\mathcal{N}(\mu_1, \Sigma_1) \big) \\
    & = \frac{1}{2}\Big[ \log\frac{\det(\Sigma_1)}{\det(\Sigma_1+\sigma^2\mathbf{I})} -d + \Tr\big(\Sigma_1^{-1}(\Sigma_1+\sigma^2\mathbf{I})\big) \Big] \\
    & = \frac{1}{2}\Big[ \log \det \Sigma_1 - \log\det (\Sigma_1 + \sigma^2\mathbf{I}) + \sigma^2\Tr(\Sigma_1^{-1}) \Big] \\
    & \explainEq{i} \frac{1}{2}\Big[ \sum_{j=1}^d \sigma_{(1,j)}^2 - \sum_{j=1}^d (\sigma_{(1,j)}^2 +\sigma^2) + \sigma^2 \sum_{j=1}^d \frac{1}{\sigma_{(1,j)}^2} \Big]
    =\frac{\sigma^2}{2}\sum_{j=1}^d(\frac{1}{\sigma_{(1,j)}^2}-1), \label{supp:KL:div:additive:Gauss}
\end{align}
where (i) assumes $\Sigma_1$ follows the diagonal structure $\Sigma_1 = \diag[\sigma_{(1,1)}^2, \dots, \sigma_{(1,j)}^2, \dots, \sigma_{(1,d)}^2]$ as it does in the VAE setting. Since $\sigma^2 \geq \frac{2 L^2  \log \frac{1.25}{\delta} }{\varepsilon^2}$ (by \thmref{supp:thm:gaussian:mech:dp}) guarantees the Gaussian mechanism is differentially private, we can proof that the KL-distortion budget allows for differential privacy through the following proposition:
\begin{proposition}{}
The additive Gaussian mechanism with the adjustable variance $\sigma^2$ is $(\varepsilon, \delta)$-differentially private, when the distortion budget $b$ satisfies 
\begin{align}
    \frac{L^2\log\frac{1.25}{\delta}}{\varepsilon^2}\Big( \sum_{j=1}^d\big(\frac{1}{\sigma_{(1,j)}^2}-1\big) \Big) \leq b,  
\end{align}
where $L = \max_{z\simeq z'}||z - z'||_2, \forall z, z' \in \mathcal{Z}$, and $\sigma_{(1,j)}^2$ is the diagonal elements of the sample covariance $\Sigma_1$.
\end{proposition}  
The proof of this proposition immediately follows from using \thmref{supp:thm:gaussian:mech:dp} on \eqrefp{supp:KL:div:additive:Gauss}.  
This relationship is intuitive as increasing the budget of divergence $b$ will implicitly allow larger variance of the additive noise $w$. Thus it becomes difficult to distinguish whether $\tilde{z}$ originated from $z$ or $z'$ in latent space.
The above example is just one complementary case of how the basic Gaussian mechanism falls into our divergence constraints in the context of differential privacy. There are many other potential extensions such as obfuscating the gradient, or a randomized response to generate new samples.

In our implementation, the noise $w$ is generated from $\mathcal{N}(0, \sigma^2\mathbf{I})$ where $\sigma^2$ is learned through the back propagation of the robust optimization loss function subject to $D_{\text{KL}}(q_{\psi}||q_{\phi})\leq b$ with a prescribed $b$.

\subsection{Why does cross-entropy loss work}
In this section, we build the connection between cross-entropy loss and mutual information to give intuition for why maximizing the cross-entropy loss in our optimization reduces the correlation between released data and sensitive labels. Given that the encoder part of the VAE is fixed, we focus on the latent variable $z$ for the remaining discussion in this section.   

The mutual information between latent variable $z$ and sensitive label $y$ can be expressed as follows  
\begin{align}
    I(z;y) & =\mathbb{E}_{q(z, y)}\Big[\log q(y|z) - \log q(y) \Big] \\
    & = \mathbb{E}_{q(z, y)}\big[ \log q(y|z) - \log p(y|z) - \log q(y) + \log p(y|z) \big] \\
    & = \mathbb{E}_{q(y|z)q(z)}\big[ \log \frac{q(y|z)}{p(y|z)} \big] + \mathbb{E}_{q(z,y)} [\log p(y|z) ] - \mathbb{E}_{q(y|z)q(y)}[\log q(y)] \\
    & = \underbrace{\mathbb{E}_{q(z)}\Big[ D_{KL}\big(q(y|z)|| p(y|z)\big) \Big]}_{\geq 0} + \mathbb{E}_{q(z, y)}[\log p(y|z)] + H(y) \\
    & \geq \mathbb{E}_{q(z, y)}[\log p(y|z)] + H(y), \label{supp:eq:I:vari:low:bd}
\end{align}
where $q$ is the data distribution, and $p$ is the approximated distribution. which is similar to the last logit layer of a neural network classifier. Then, the term $-\mathbb{E}_{q(z, y)}[\log p(y|z)]$ is the cross-entropy loss $H(q, p) $ (The corresponding negative log-likelihood is $-\mathbb{E}_{q(z, y)}[\log p(y|z)]$ ). In classification problems, minimizing the cross-entropy loss enlarges the value of $\mathbb{E}_{q(z, y)}[\log p(y|z)]$. Consequently, this pushes the lower bound of $I(z; y)$ in \eqrefp{supp:eq:I:vari:low:bd} as high as possible, indicating high mutual information. 

However, in our robust optimization, we maximize the cross-entropy, thus, decreasing the value of $\mathbb{E}_{q(z, y)}[\log p(y|z)]$ (More specifically, it is $\mathbb{E}_{q(\tilde{z}, y)}[\log p(y|\tilde{z})]$, given the mutual information we care about is between the new representation $\tilde{z}$ and sensitive label $y$ in our application). Thus, the bound of \eqrefp{supp:eq:I:vari:low:bd} has a lower value which indicates the mutual information $I(\tilde{z}; y)$ can be lower than before. Such observations can also be supported by the empirical results of mutual information shown in \figref{fig:mnist:mi:plot:case1:case2}.  
\subsection{Experiment Details}
\label{supp:nn:archi:explain}
\subsubsection{VAE architecture}
The MNIST dataset contains 60000 samples of gray-scale handwritten digits with size 28-by-28 pixels in the training set, and 10000 samples in the testing set.  
When running experiments on MNIST, we convert the $28\times 28$ images into 784 dimensional vectors, and construct a network with the following structure for the VAE:
\begin{align*}
    & x \rightarrow \text{FC}(300) \rightarrow \text{ELU} \rightarrow \text{FC}(300) \rightarrow \text{ELU} \rightarrow \text{FC}(20) (\text{split $\mu$ and $\Sigma$ to approximate $q(z|x)$}) \\
    & z \rightarrow \text{FC}(300) \rightarrow \text{ELU} \rightarrow \text{FC}(300) \rightarrow \text{ELU} \rightarrow \text{FC}(784).
\end{align*}

The aligned CelebA dataset contains 202599 samples. We crop each image down to 64-by-64 pixels with 3 color (RGB) channels and pick the first 182000 samples as the training set, and leave the remainder as the testing set. The encoder and decoder architecture for CelebA experiments are described in Table~\ref{tab:encode:archi} and Table~\ref{tab:decode:archi}.

We use Adam \citep{kingma2014adam} to optimize the network parameters throughout all training procedures, with a batch size equal 100 and 24 for MNIST and CelebA respectively. 
\begin{table}[!hpbt]
\vspace{-0.1in}
    \caption{Encoder Architecture in CelebA experiments. We use the DenseNet \citep{huang2017densely} architecture with a slight modification to embed the raw images into a compact latent vector, with growth rate $m=12$ and depth $k=82$}
    \label{tab:encode:archi}
    \centering
    \begin{small}
    \begin{tabular}{c|c|c}
    \toprule
         \textbf{Name} & \textbf{Configuration} & \textbf{ Replication} \\
    \midrule
         \multirow{2}{*}{initial layer} & conv2d=(3, 3), stride=(1, 1), & \multirow{2}{*}{1} \\
         & padding=(1, 1), channel in = 3, channel out = $2m$ &  \\
    \midrule 
         \multirow{3}{*}{dense block1} & batch norm, relu, conv2d=(1, 1), stride=(1, 1), & \\
         & batch norm, relu, conv2d=(3, 3), stride=(1, 1), & 12 \\
         & growth rate = $m$, channel in = 2$m$  & \\ 
    \midrule      
        \multirow{4}{*}{transition block1} &  batch norm, relu, & \multirow{4}{*}{1} \\
        & conv2d=(1, 1), stride=(1, 1), average pooling=(2, 2), &  \\
        & channel in = $ \frac{(k-4)}{6} m + 2m$, & \\
        & channel out = $\frac{(k-4)}{12} m + m$ &  \\
    \midrule 
    \multirow{3}{*}{dense block2} & batch norm, relu, conv2d=(1, 1), stride=(1, 1), & \multirow{3}{*}{12} \\
         & batch norm, relu, conv2d=(3, 3), stride=(1, 1), & \\
         & growth rate=$m$, channel in = $ \frac{(k-4)}{12} m + m$, & \\
    \midrule      
        \multirow{4}{*}{transition block2} &  batch norm, relu, & \multirow{4}{*}{1}\\
        & conv2d=(1, 1), stride=(1, 1), average pooling=(2, 2), & \\
        & channel in = $\frac{(k-4)}{6} m  + \frac{(k-4)}{12} m + m $ &  \\ 
        & channel out= $\frac{1}{2}\big(\frac{(k-4)}{6} m  + \frac{(k-4)}{12} m + m \big)$ & \\
    \midrule 
    \multirow{3}{*}{dense block3} & batch norm, relu, conv2d=(1, 1), stride=(1, 1), & \multirow{3}{*}{ 12} \\
         & batch norm, relu, conv2d=(3, 3), stride=(1, 1), & \\ 
         & growth rate = $m$, channel in = $\frac{1}{2}\big(\frac{(k-4)}{6} m  + \frac{(k-4)}{12} m + m \big)$ & \\ 
    \midrule      
        \multirow{4}{*}{transition block3} &  batch norm, relu, & \multirow{4}{*}{1} \\ 
        & conv2d=(1, 1), stride=(1, 1), average pooling=(2, 2), \\
        & channel in = $\frac{1}{2}\big(\frac{(k-4)}{6} m  + \frac{(k-4)}{12} m + m \big) +\frac{(k-4)}{6}m $  &  \\
        & channel out = $\frac{1}{2}\big(\frac{1}{2}\big(\frac{(k-4)}{6} m  + \frac{(k-4)}{12} m + m \big) +\frac{(k-4)}{6}m \big)$ & \\
    \midrule 
        output layer & batch norm, fully connected 100  & 1 \\
    \bottomrule     
    \end{tabular}
    \end{small}
\end{table}

\begin{table}[!hpbt]
\vspace{-0.0in}
    \caption{Decoder Architecture in CelebA experiments. }
    \label{tab:decode:archi}
    \centering
    \begin{small}
    \begin{tabular}{c|c|c}
    \toprule
    \textbf{Name} & \textbf{Configuration} & \textbf{ Replication} \\
    \midrule
    initial layer & fully connected 4096 & 1 \\
    \midrule
    reshape block & resize 4096 to $256 \times 4 \times 4$ & 1 \\
    \midrule 
    \multirow{3}{*}{deccode block} & conv transpose=(3, 3), stride=(2, 2), & \multirow{3}{*}{4}\\
    & padding=(1, 1), outpadding=(1, 1), & \\
    & relu, batch norm & \\
    \midrule 
    \multirow{2}{*}{decoder block} & conv transpose=(5, 5), stride=(1, 1), & \multirow{2}{*}{1}\\
    & padding=(2, 2) &  \\
    \midrule 
    \end{tabular}
    \end{small}
\end{table}

\subsubsection{Filter Architecture}
We use a generative linear filter throughout our experiments. In the MNIST experiments, we compressed the latent embedding down to a 10-dim vector. For \textbf{MNIST Case 1}, we use a 10-dim Gaussian random vector $\epsilon$ concatenated with a 10-dim one-hot vector $y$ representing digit id labels, where $\epsilon \sim \mathcal{N}(0, I)$ and $y \in \{0 ,1\}^{10}$. We use the linear filter $A$ to ingest the concatenated vector, and add the corresponding output to the original embedding vector $z$ to yield $\tilde{z}$. Thus the mechanism is
\begin{align}
    \tilde{z}=f(z, \epsilon, y) = z + A \begin{bmatrix}\epsilon \\ y\end{bmatrix} \label{appx:filter:scheme},
\end{align}
where $A \in \mathbb{R}^{10\times 20}$ is a matrix. For \textbf{MNIST Case 2}, we use a similar procedure except the private label $y$ is a binary label (i.e. digit value $\geq 5$ or not). Thus, the corresponding one-hot vector is 2-dimensional. As we keep $\epsilon$ to be a 10-dimensional vector, the corresponding linear filter $A$ is a matrix in $\mathbb{R}^{10 \times 12}$. 

In the experiment of CelebA, we create the generative filter following the same methodology in \eqrefp{appx:filter:scheme}, with some changes on the dimensions of $\epsilon$ and $A$. (i.e. $\epsilon \in R^{50}$ and $A \in \mathbb{R}^{50 \times 52}$)   
%
%
%
\subsubsection{Adversarial classifiers}
In the MNIST experiments, we use a small architecture consisting of neural networks with two fully-connected layers and an exponential linear unit (ELU) to serve as the privacy classifiers, respectively. The specific structure of the classifier is depicted as follows:
\begin{align*}
    & z \text{ or } \tilde{z} \rightarrow \text{FC}(15) \rightarrow \text{ELU} \rightarrow \text{FC}(y) [\text{ or FC}(u) ].
\end{align*} 
In the CelebA experiments, we construct a two-layered neural network that is shown as follows:
\begin{align*}
    & z \text{ or } \tilde{z} \rightarrow \text{FC}(60) \rightarrow \text{ELU} \rightarrow \text{FC}(y) [\text{ or FC}(u) ].
\end{align*} 
The classifiers ingest the embedded vectors and output unnormalized logits for the private label or utility label. The classification results can be found in table~\ref{tab:celebA:clf:acc:results}.  

\subsection{More results of MNIST experiments}
\label{supp:mnist:more:res:w:mi}
\begin{figure}[!hpbt]
    \centering
    \includegraphics[width=0.51\textwidth]{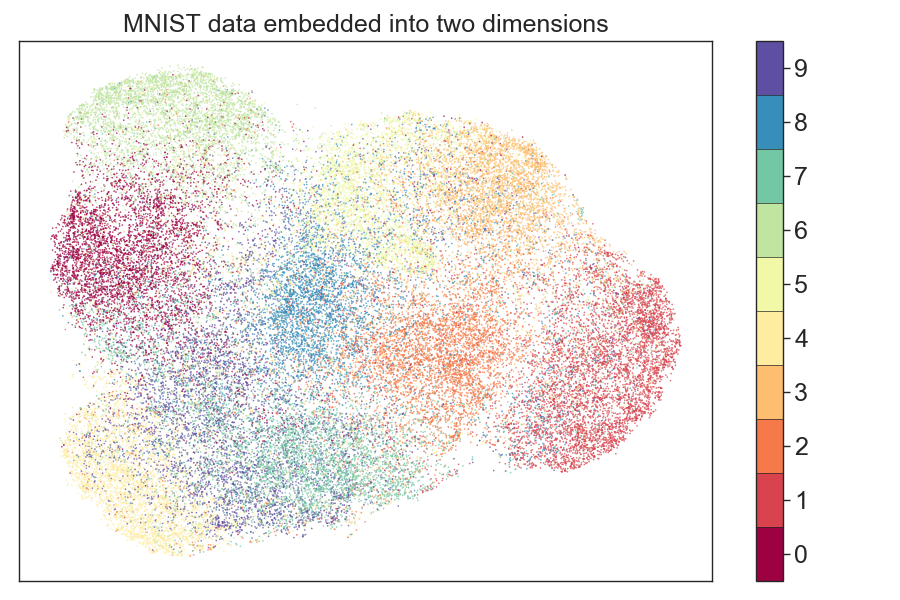}
    \caption{\textbf{MNIST case 2}: Visualization of the perturbed latent geometry. We discover that \textsf{0} is closer to \textsf{6}, compared with the original latent geometry in \figref{fig:circle:latent:geo:raw}, which indicates that it would be more difficult to distinguish which of those two digits is larger than or equal to five, even though both are even digits. }
    \label{fig:mnist:latent:geo:oddeven}
\end{figure}

In this section, we illustrate detailed results for the MNIST experiment when we set whether the digit is odd or even as the utility label, and the whether the digit is greater than or equal to 5 as the private label. We first show samples of raw images and privatized images in \figref{fig:MNIST:samples:odd:even}. We show the classification accuracy and its sensitivity in \figref{fig:MNIST:clf:acc:odd:even}. Furthermore, we display the geometry of the latent space in \figref{fig:mnist:latent:geo:oddeven}. 

In addition to the classification accuracy, we evaluate the mutual information, to justify our generative filter indeed decreases the correlation between released data and private labels, as shown in \figref{fig:mnist:mi:plot:case1:case2}.

\subsubsection{Utility of Odd or Even}
\begin{figure*}[!hpbt]
    \centerline{
    \begin{subfigure}[t]{0.51\columnwidth}
    \includegraphics[width=1\textwidth]{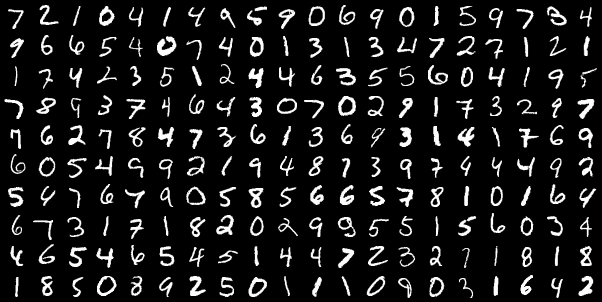}
    \caption{Sample of original digits}
    \end{subfigure}
    \begin{subfigure}[t]{0.51\columnwidth}
    \includegraphics[width=1\textwidth]{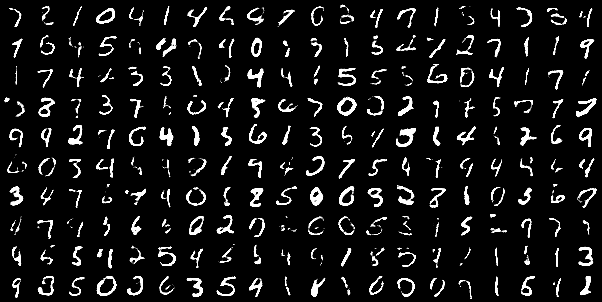}
    \caption{Same images large-valued digits privatized}
    \end{subfigure}
    }
    \caption{\textbf{MNIST case 2}: Visualization of digits pre and post-noise injection and adversarial training. We discover that some large-valued digits ($\geq 5$) are randomly switched to low-valued ($<5$) digits (or vice versa) while some even digits remain even digits and odd digits remain as odd digits. }
    \label{fig:MNIST:samples:odd:even}
\end{figure*}

\begin{figure*}[!hpbt]
    \centerline{
    \begin{subfigure}[t]{0.37\columnwidth}
    \includegraphics[width=1\textwidth]{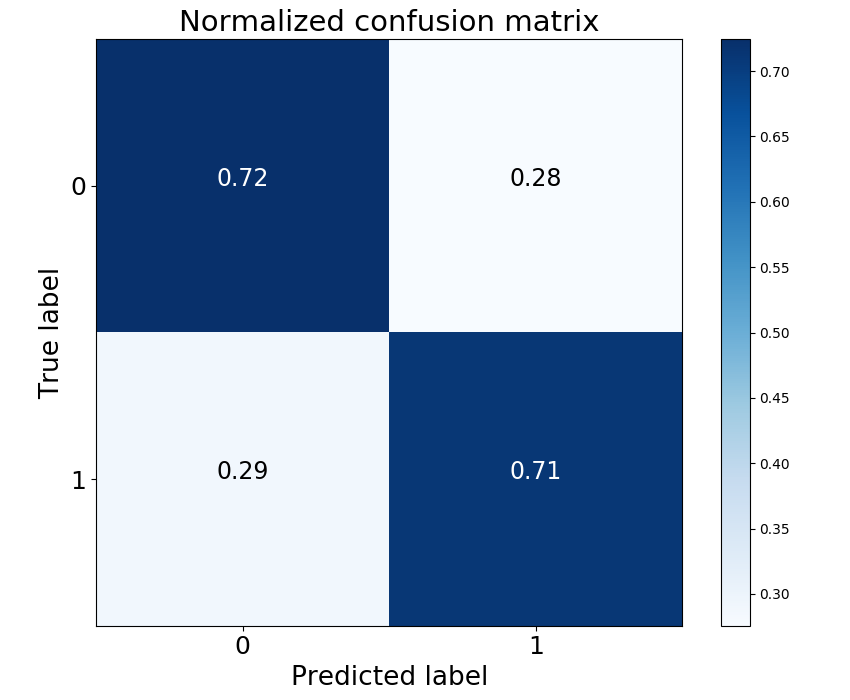}
    \caption{Confusion matrix of classifying the private label when $b=3$}
    \label{fig:mnist:confu:mat:priv:geq5:util:oddeven}
    \end{subfigure}
    ~
    \begin{subfigure}[t]{0.41\columnwidth}
    \includegraphics[width=1\textwidth]{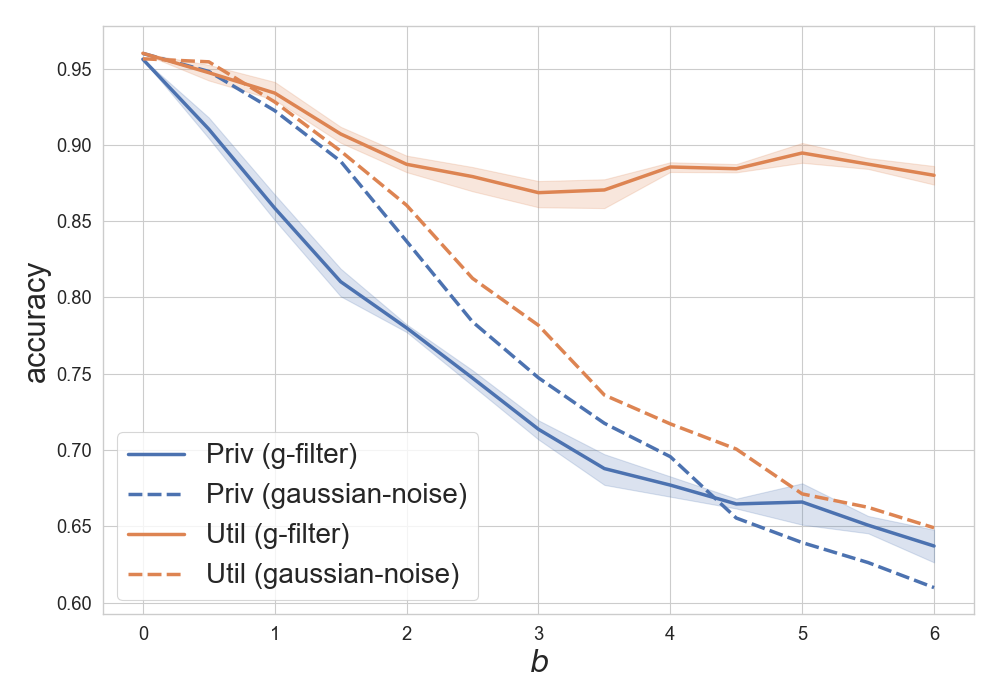}
    \caption{Sensitivity of classification accuracy versus the distortion $b$}
    \label{fig:mnist:acc:inc:b:priv:geq5:util:oddeven}
    \end{subfigure}
    ~
    \begin{subfigure}[t]{0.41\columnwidth}
    \includegraphics[width=1\textwidth]{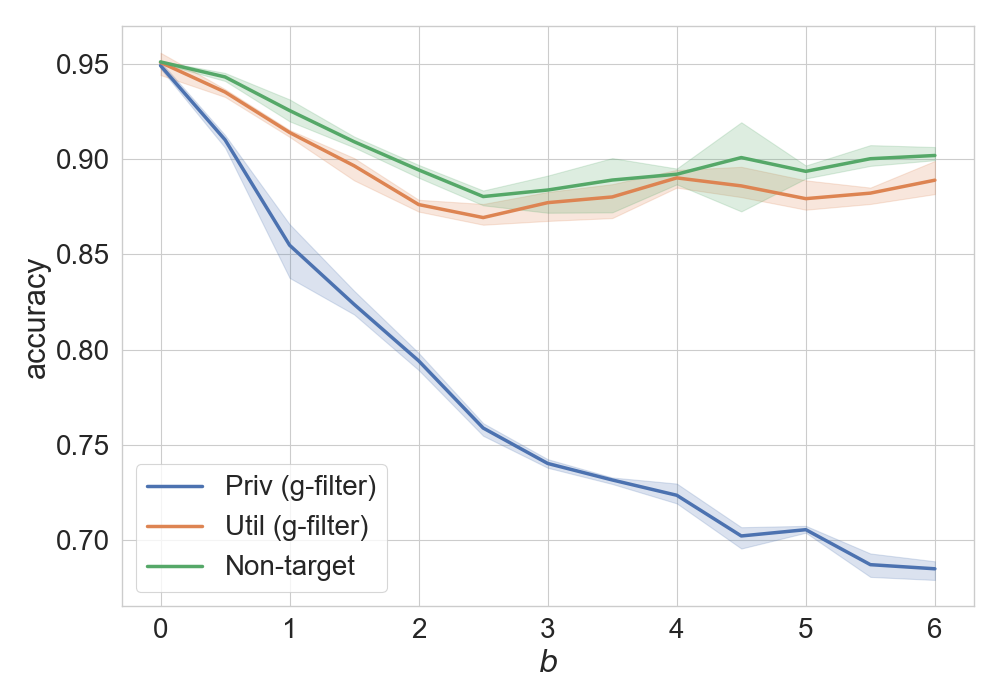}
    \caption{Sensitivity of classification accuracy versus the distortion $b$ with a non-target attribute}
    \label{fig:mnist:acc:inc:b:priv:geq5:util:non-target}
    \end{subfigure}
    }
    \caption{\textbf{MNIST case 2}: \figref{fig:mnist:confu:mat:priv:geq5:util:oddeven} shows the false positive and false negative rates for classifying the private label when the distortion budget is 3 in KL-divergence. The \figref{fig:mnist:acc:inc:b:priv:geq5:util:oddeven} shows that when we use the generative adversarial filter, the classification accuracy of private labels drops from 95\% to almost 65\% as the distortion increases, while the utility label can still maintain close to 90\% accuracy throughout. Meanwhile, the additive Gaussian noise performs worse because it yields higher accuracy on the private label and lower accuracy on the utility label, compared to the generative adversarial filter. \Figref{fig:mnist:acc:inc:b:priv:geq5:util:non-target} shows how non-target attributes not included in the filter's loss function (circle) can still be preserved due to the distortion budget restricting noise injection.}
    \label{fig:MNIST:clf:acc:odd:even}
\end{figure*}

We present some examples of digits when the utility is odd or even number in \figref{fig:MNIST:samples:odd:even}. The confusion matrix in \figref{fig:mnist:confu:mat:priv:geq5:util:oddeven} shows that false positive rate and false negative rate are almost equivalent, indicating the perturbation resulted from filter doesn't necessarily favor one type (pure positive or negative) of samples. \Figref{fig:mnist:acc:inc:b:priv:geq5:util:oddeven} shows that the generative filter, learned through minmax robust optimization, outperforms the Gaussian mechanism under the same distortion budget. The Gaussian mechanism reduces the accuracy of both private and utility labels, while the generative filter can maintain the accuracy of the utility while decreasing the accuracy of the private label, as the distortion budget goes up.

Furthermore, the distortion budget prevents the generative filter from distorting non-target attributes too severely. This allows the data to maintain some information even if it is not specified in the filter's loss function. \Figref{fig:mnist:acc:inc:b:priv:geq5:util:non-target} shows the accuracy with the added non-target label of circle from MNIST case 1. 

\subsubsection{Empirical Mutual information}
\begin{figure}[!hbpt]
    \centerline{
    \begin{subfigure}[t]{0.45\columnwidth}
    \includegraphics[width=1\textwidth]{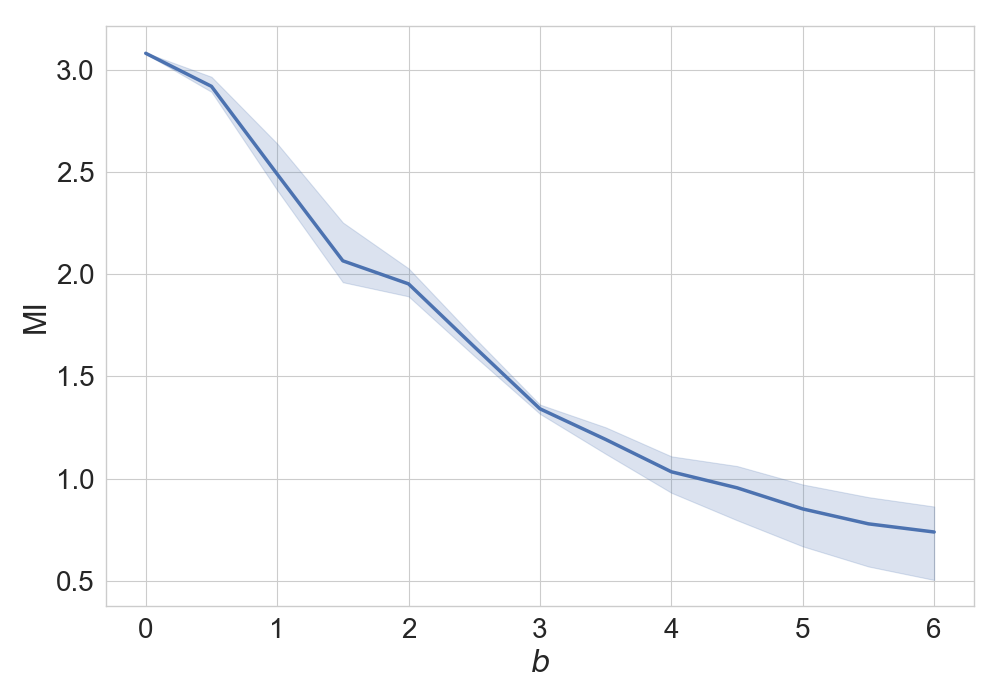}
    \caption{Digit identity as the private label}
    \end{subfigure}
    \begin{subfigure}[t]{0.45\columnwidth}
    \includegraphics[width=1\textwidth]{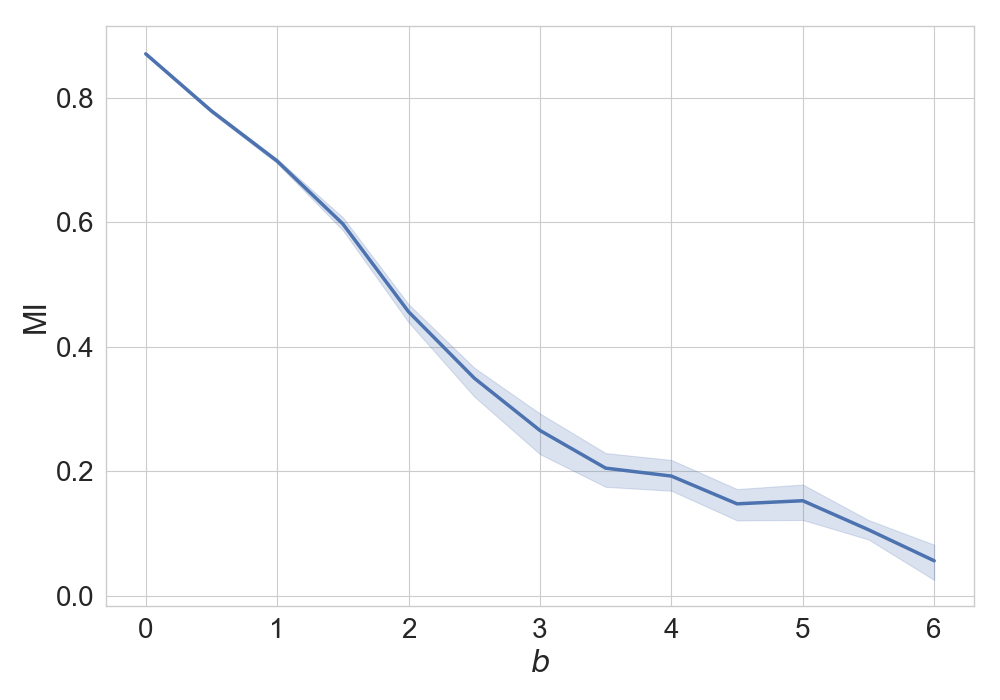}
    \caption{Large- or small-value as the private label}
    \end{subfigure}
    }
    \caption{Mutual information between the perturbed embedding and the private label decreases as the distortion budget grows, for both MNIST case 1 and case 2. }
    \label{fig:mnist:mi:plot:case1:case2}
\end{figure}

We leverage the empirical mutual information \citep{gao2015efficient} to verify if our new perturbed data is less correlated with the sensitive labels from an  information-theoretic perspective. The empirical mutual information is clearly decreased as shown in  \Figref{fig:mnist:mi:plot:case1:case2}, which supports that our generative adversarial filter can protect the private label given a certain distortion budget. 

%
%
\subsection{More examples of CelebA}
\label{appx:sec:celebA:examples:more}
\begin{figure*}[!hpbt]
    \centerline{
    \begin{subfigure}[t]{0.54\columnwidth}
    \includegraphics[width=1\textwidth]{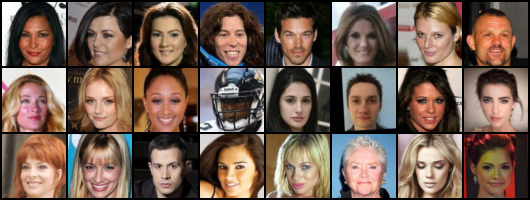}
    \end{subfigure}
    \begin{subfigure}[t]{0.54\columnwidth}
    \includegraphics[width=1\textwidth]{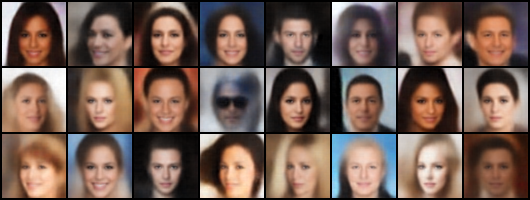}
    \end{subfigure}
    }
    \caption{Visualizing raw samples (on left) and encoded-decoded samples (on right) from a trained VAE with the Lipschitz smoothness.}
    \label{fig:celebA:raw:enc:dec:samples}
\end{figure*}

\begin{figure*}[!hpbt]
    \centerline{
    \begin{subfigure}[t]{0.54\columnwidth}
    \includegraphics[width=1\textwidth]{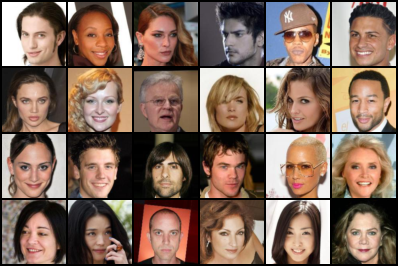}
    \caption{raw}
    \label{fig:celebA:samples:raw:attractive}
    \end{subfigure}
    \begin{subfigure}[t]{0.54\columnwidth}
    \includegraphics[width=1\textwidth]{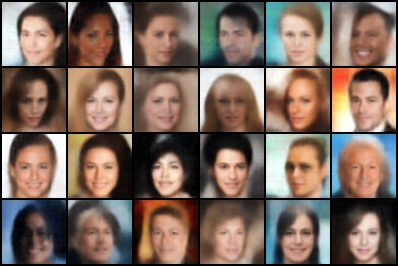}
    \caption{attractive privatized}
    \label{fig:celebA:samples:priv:attractive}
    \end{subfigure}
    }
    \caption{Sampled images. We find some non-attractive faces switch to attractive faces and some attractive looking images are changed into non-attractive, from left \figref{fig:celebA:samples:raw:attractive} to right \figref{fig:celebA:samples:priv:attractive}. }
    \label{fig:celebA:example:priv:attractive}
\end{figure*}

\begin{figure*}[!hpbt]
    \centerline{
    \begin{subfigure}[t]{0.54\columnwidth}
    \includegraphics[width=1\textwidth]{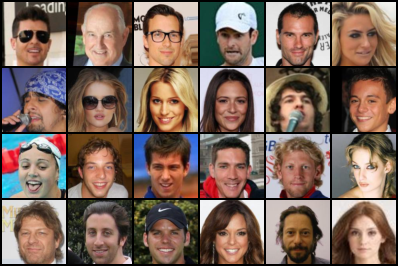}
    \caption{raw}
    \label{fig:celebA:samples:raw:eyegls}
    \end{subfigure}
    \begin{subfigure}[t]{0.54\columnwidth}
    \includegraphics[width=1\textwidth]{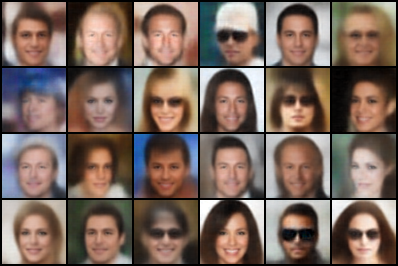}
    \caption{eyeglass privatized}
    \label{fig:celebA:samples:priv:eyegls}
    \end{subfigure}
    }
    \caption{Sampled images. We find some faces with eyeglasses are switched to non-eyeglasses faces and some non-eyeglasses faces are changed into eyeglasses-wearing faces, from left \figref{fig:celebA:samples:raw:eyegls} to right \figref{fig:celebA:samples:priv:eyegls}. }
    \label{fig:celebA:example:priv:eyeglasses}
\end{figure*}

\begin{figure*}[!hpbt]
    \centerline{
    \begin{subfigure}[t]{0.54\columnwidth}
    \includegraphics[width=1\textwidth]{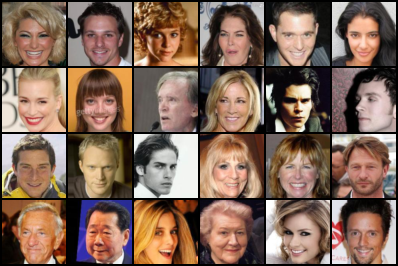}
    \caption{raw}
    \label{fig:celebA:samples:raw:wavehair}
    \end{subfigure}
    \begin{subfigure}[t]{0.54\columnwidth}
    \includegraphics[width=1\textwidth]{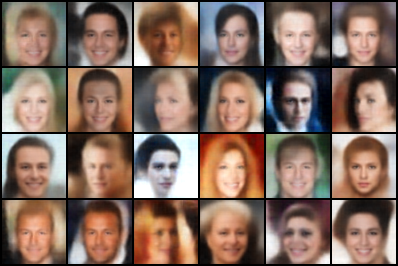}
    \caption{wavy hair privatized}
    \label{fig:celebA:samples:priv:wavehair}
    \end{subfigure}
    }
    \caption{Sampled images. We discover that some faces with wavy hair switch to images with non-wavy hair after our filter's perturbation (and vice versa), from left \figref{fig:celebA:samples:raw:wavehair} to right \figref{fig:celebA:samples:priv:wavehair}. }
    \label{fig:celebA:example:priv:wavyhair}
\end{figure*}

In this part, we illustrate more examples of CelebA faces yielded by our generative adversarial filter (Figures \ref{fig:celebA:example:priv:attractive}, \ref{fig:celebA:example:priv:eyeglasses}, and \ref{fig:celebA:example:priv:wavyhair}). We show realistic looking faces generated to privatize the following labels: attractive, eyeglasses, and wavy hair, while maintaining smiling as the utility label. The blurriness of the images is typical of state of the art VAE models due to the compactness of the latent representation \citep{higgins2017beta, dupont2018learning}. The blurriness is not caused by the privatization procedure, but by the encoding and decoding steps as demonstrated in \figref{fig:celebA:raw:enc:dec:samples}.

\end{document}